%% file: main.tex
\documentclass[10pt,twocolumn,letterpaper]{article}

\usepackage[pagenumbers]{cvpr} 

\usepackage{multirow}
\usepackage{array}
\usepackage{makecell}   
\usepackage{graphicx}
\usepackage{amsfonts}   
\usepackage{amsmath}    
\usepackage{cite}      
\usepackage{booktabs}   
\usepackage[table,xcdraw]{xcolor}   
\usepackage{bm}         
\usepackage{color}      
\usepackage{algorithm}
\usepackage{algorithmic}
\usepackage{multirow}
\usepackage{tikz}
\usepackage{pgfplots}
\usepackage{amssymb}
\usepackage{adjustbox}
\usepackage{amsmath}
\usepackage{bbm}

\definecolor{cvprblue}{rgb}{0.21,0.49,0.74}
\usepackage[pagebackref,breaklinks,colorlinks,citecolor=cvprblue]{hyperref}

\def\paperID{*****} 
\def\confName{CVPR}
\def\confYear{2025}

\title{CLIP-GS: CLIP-Informed Gaussian Splatting for View-Consistent 3D Indoor Semantic Understanding}

\author{Guibiao Liao$^{1,2}$ \quad Jiankun Li$^{3}$ \quad Zhenyu Bao$^{1,2}$ \quad Xiaoqing Ye$^{3}$ \quad Qing Li$^{2}$ \quad Kanglin Liu$^{2}$\thanks{Corresponding author} \\ 
\textsuperscript{\rm 1}Peking University,
\textsuperscript{\rm 2}Pengcheng Laboratory, 
\textsuperscript{\rm 3}Baidu Inc. 
}

\begin{document}
\maketitle
\input{sec/0_abstract}

\input{sec/1_mainpaper}

{
    \small
    \bibliographystyle{ieeenat_fullname}
    \bibliography{main}
}


\end{document}

%% file: sec/0_abstract.tex
\begin{abstract}
Exploiting 3D Gaussian Splatting (3DGS) with Contrastive Language-Image Pre-Training (CLIP) models for open-vocabulary 3D semantic understanding of indoor scenes has emerged as an attractive research focus. Existing methods typically attach high-dimensional CLIP semantic embeddings to 3D Gaussians and leverage view-inconsistent 2D CLIP semantics as Gaussian supervision, resulting in efficiency bottlenecks and deficient 3D semantic consistency. 
To address these challenges, we present CLIP-GS, efficiently achieving a coherent semantic understanding of 3D indoor scenes via the proposed Semantic Attribute Compactness (SAC) and 3D Coherent Regularization (3DCR). 
SAC approach exploits the naturally unified semantics within objects to learn compact, yet effective, semantic Gaussian representations, enabling highly efficient rendering (\textgreater100 FPS). 
3DCR enforces semantic consistency in 2D and 3D domains: In 2D, 3DCR utilizes refined view-consistent semantic outcomes derived from 3DGS to establish cross-view coherence constraints; in 3D, 3DCR encourages features similar among 3D Gaussian primitives associated with the same object, leading to more precise and coherent segmentation results. 
Extensive experimental results demonstrate that our method remarkably suppresses existing state-of-the-art approaches, achieving mIoU improvements of 21.20\% and 13.05\% on ScanNet and Replica datasets, respectively, while maintaining real-time rendering speed. 
Furthermore, our approach exhibits superior performance even with sparse input data, substantiating its robustness. 
\end{abstract}

%% file: sec/1_mainpaper.tex
\begin{figure*}[!t]
\centering
\includegraphics[width=\textwidth]{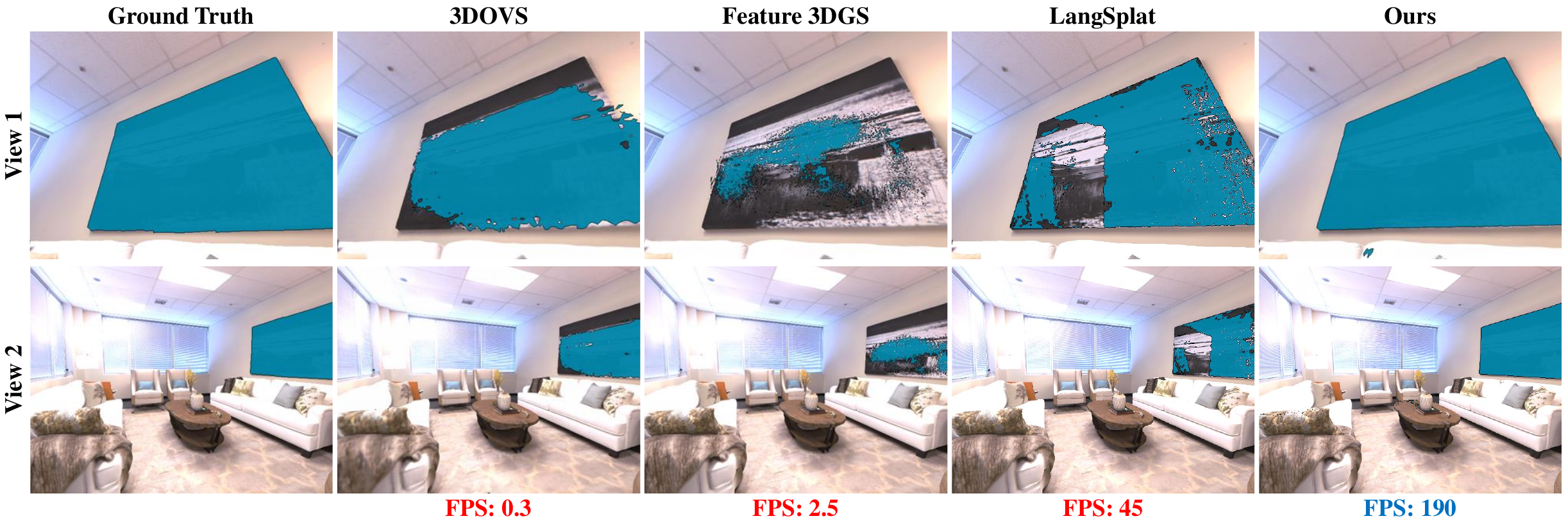}
\caption{
Visual comparisons between different CLIP-informed 3D semantic segmentation methods using the text query \underline{"Picture"} across different views. 
The \textit{NeRF}-based method, 3DOVS \cite{3DOVS}, and \textit{3DGS}-based methods, Feature 3DGS \cite{feature3dgs} and LangSplat \cite{langsplat} exhibit ambiguous semantics and efficiency bottlenecks. In contrast, our approach achieves more precise and consistent semantic segmentation results with a faster speed. 
} 
\label{fig:vis_fig1}
\end{figure*}

\section{Introduction} \label{sec:intro} 
Neural Radiance Fields (NeRFs) \cite{nerf} and 3D Gaussian Splatting (3DGS) \cite{3dgs} have emerged as promising methods for high-quality 3D scene modeling and novel view synthesis \cite{nerfreview,liu2023novel,liu2022spatial,zhao2024exploiting}. Recent methods have made remarkable advancements in rendering novel views that encompass geometric and appearance details \cite{zipnerf, mipsplatting}. However, achieving a comprehensive semantic understanding of 3D scenes \cite{robotnavigation, liao2024rethinking, liao2022cross, liao2020mmnet, liao2025spc, zhang2022tdrnet} remains a challenging task. 
To achieve it, one intuitive approach involves using manually annotated semantic labels to offer semantic supervision for existing 3D scene representations. 
Nevertheless, this resource-intensive manual annotation process impedes its practical application in real-world 3D semantic understanding. 
Compared to the traditional semantic labeling manner, the 2D vision-language model, Contrastive Language-Image Pre-Training (CLIP) is exploited to provide a new approach to semantic understanding without reliance on annotated image labels. 
The CLIP model, comprising an image encoder and a text encoder, is pre-trained on extensive image-text pairs collected from websites to establish vision-language associations. This paradigm enables CLIP to exhibit promising open-vocabulary semantic understanding capabilities, allowing it to segment objects based on textual queries \cite{LSeg, CLIPSelf, vlm2scene, ovnerf, zhang2023simple}. 
Consequently, effectively harnessing image-text knowledge from CLIP for precise open-vocabulary 3D semantic understanding of indoor scenes is emerging as a valuable area.

Recently, LERF \cite{LERF} pioneers a NeRF-based semantic field optimization with CLIP visual features, enabling text-driven 3D segmentation. Building upon this foundation, 3DOVS \cite{3DOVS} further introduces a Relevancy-Distribution Alignment loss for segmentation accuracy improvement. 
However, NeRF's ray-marching volume rendering technique significantly impedes rendering efficiency (as illustrated in the 2$^{nd}$ column of Fig. \ref{fig:vis_fig1}).  
To handle this computational limitation, several methods attempt to employ 3D Gaussian representation and tile-based rasterization for rendering acceleration.  
For example, Feature 3DGS \cite{feature3dgs} embeds high-dimensional semantic parameters into 3D Gaussians and optimizes them with CLIP semantic features.  
The state-of-the-art LangSplat \cite{langsplat} learns low-dimensional, compressed CLIP features at 3D Gaussians to accelerate rasterization, and then utilizes a pre-trained deep neural network for post-process feature upsampling to obtain semantic representation. Moreover, LangSplat utilizes region masks derived from the Segment Anything Model (SAM) \cite{sam} to refine CLIP feature extraction, enhancing the precision of object boundaries for each training view.

Despite recent advances, certain limitations remain in achieving efficient and view-consistent 3D semantic understanding. 
\textbf{1) Efficiency Bottleneck.} 
While adopting efficient rasterization techniques, the high-dimensional feature rasterization in Feature 3DGS and the post-process feature upsampling in LangSplat make them fall short of achieving extremely efficient rendering. 
\textbf{2) Deficient 3D Semantic Consistency.} 
Additionally, although LangSplat employs SAM's masks to refine object boundaries in each single training view, it struggles to achieve view-consistent semantic results as illustrated in Fig. \ref{fig:vis_fig1}. 
This limitation stems from the application of view-inconsistent 2D CLIP semantics \cite{CLIPSelf} for 3D Gaussians optimization, without incorporating effective 3D-consistent constraints for cross-view consistency enhancement.

To tackle these limitations, we present \textbf{CLIP-GS}, an efficient method for achieving precise 3D semantic understanding using CLIP-informed 3D Gaussian representation. 
\textbf{To address the efficiency challenge}, we introduce the Semantic Attribute Compactness (SAC) approach. 
Motivated by the naturally unified semantics within objects, 
{
SAC extracts a single representative semantic feature for each object using SAM masks \cite{sam}, and uses low-dimensional semantic indices to encapsulate them.  
In virtue of this low-dimensional representation, we attach low-dimensional semantic embeddings to 3D Gaussians for rasterization, 
}
extremely enhancing semantic rendering efficiency in 3D scenes (\textgreater100 FPS). 
\textbf{To tackle deficient 3D consistency}, we propose 3D Coherent Regularization (3DCR) with 2D and 3D-level consistency constraints. 
Specifically, in the 2D view space, 3DCR leverages enhanced inherently view-consistent rendered outcomes from 3D models (i.e., trained 3D Gaussians) as coherent supervision signals, providing view-consistent semantic constraints. 
Additionally, in the 3D point space, 3DCR identifies 3D Gaussian primitives associated with the same object via ray-based intersection matching, and explicitly encourages semantic similarity among their semantic embeddings to enhance semantic consistency. 
By incorporating this semantic-consistent optimization, our method achieves view-consistent segmentation results, as shown in the last column of Fig. \ref{fig:vis_fig1}.

To evaluate the 3D semantic understanding performance of our approach, we conduct experiments on synthetic and real-world datasets: Replica \cite{replica}, ScanNet \cite{scannet}, and 3DOVS \cite{3DOVS}. 
Evaluations demonstrate superior performance of our method compared to existing state-of-the-art methods, especially achieving significant mIoU improvements of 21.20\% and 13.05\% on ScanNet and Replica, respectively. 
The key contributions of this work are as follows.

\begin{itemize}
    
    \item We introduce a Semantic Attribute Compactness (SAC) approach that efficiently attaches compact and effective semantic information into 3D Gaussians, ensuring extremely efficient rendering.  
    
    \item We propose a 3D Coherent Regularization (3DCR) approach that addresses the issue of deficient 3D consistency by imposing semantic-consistent constraints at 2D and 3D levels, leading to more coherent segmentation results across different viewpoints.  
    
    \item Extensive experiments demonstrate that our approach outperforms state-of-the-art CLIP-informed 3D semantic understanding methods in both segmentation precision and rendering efficiency (\textgreater100 FPS). 
    Moreover, our method shows superior performance in the sparse-view setting, validating its robustness. 
    
\end{itemize}

\section{Related Work}
\label{sec:related_work} 
\textbf{Gaussian Splatting for 3D Scene Representation.} 
Representing 3D scenes via radiance fields has witnessed significant advancements in recent years. 
Neural Radiance Field (NeRF) \cite{nerf} implicitly represents the appearance and geometry of 3D scenes using a coordinate-based neural network.  
Despite substantial efforts to improve optimization and rendering efficiency \cite{instantngp, tensorf}, NeRF-based methods still face challenges of slow rendering speeds, primarily due to the neural network query process and volume rendering.

Recently, Kerbl et al. \cite{3dgs} proposed 3D Gaussian Splatting (3DGS), a novel approach to represent 3D scenes as collections of 3D Gaussians. By employing a fast tile-based rasterization technique, 3DGS enables real-time rendering at 1080p resolution while maintaining high-quality visual results.
Building upon the efficiency demonstrated by 3DGS, numerous recent studies have extended its application to various tasks, such as head and human reconstruction \cite{chu2024generalizable,sun2024real,zielonka2025synshot}, autonomous driving scene modeling \cite{streetGaussian,fischer2024dynamic,zhang2024street}, and 3D editing \cite{wu2024gaussctrl,wang2024learning,in2025editsplat}. 
Besides, recent surveys \cite{fei20243d,bao20253d} provide a more comprehensive overview of the developments of 3DGS. 
Unlike these methods, our study focuses on harnessing 3DGS for CLIP-informed 3D scene semantic understanding.

\noindent
\textbf{CLIP-Informed 3D Semantic Fields.} 
Early methods focused on integrating CLIP features \cite{OpenAICLIP, OpenCLIP} into NeRFs to establish 3D semantic fields.  
DFF \cite{DFF} explored the incorporation of features from CLIP-LSeg \cite{LSeg} to optimize the semantic feature field. 
LERF \cite{LERF} extended this concept by introducing a scale-conditioned feature field supervised by multi-scale CLIP features from the CLIP visual encoder. 
Similarly, 3DOVS \cite{3DOVS} optimized the semantic feature field using CLIP features and introduced a relevance-distribution alignment loss to enhance accuracy. 
Yet, these NeRF-based methods are constrained by the computational bottleneck of NeRF's volume rendering, impeding efficient rendering.

Alternative to NeRFs, recent works explore embedding CLIP features into 3D Gaussians to construct a semantic field and employ tile-based rasterization for efficient semantic rendering. 
Feature 3DGS \cite{feature3dgs} attached high-dimensional CLIP embeddings to 3D Gaussians, optimizing them using CLIP semantic features. However, embedding high-dimensional parameters in millions of 3D Gaussians significantly hampers rendering efficiency. 
FMGS \cite{fmgs} integrated 3D Gaussians with multi-resolution hash encodings to form a combined feature field, supervised by multi-view CLIP features.  
Shi et. al. \cite{shi2023language} embedded quantized CLIP features onto 3D Gaussians to reduce memory and storage requirements. 
LangSplat \cite{langsplat} learned low-dimensional semantic Gaussian features and utilized a pre-trained deep neural network to upsample rendered features, aligning them with high-dimensional CLIP features. 
However, pre-training the deep network and applying the post-processing upsampling process inevitably introduce additional time overhead, thereby compromising the method's overall efficiency. 
Moreover, while LangSplat used SAM to generate object masks and then utilized CLIP to encode these regions for single-view object boundary refinement, it still suffers from semantic ambiguity due to a lack of cross-view, semantic-consistent supervision, resulting in limited coherence in the segmentation results. 
Note that methods for 3D interactive segmentation, such as \cite{gaugrouping, saga, flashsplat, clickgaussian}, fall outside the scope of the CLIP-informed semantic fields. These approaches primarily generate region masks devoid of semantic meaning, thereby lacking the capability to directly segment objects through language queries. 
For instance, Gaussian Grouping \cite{gaugrouping} requires supplementary techniques like Grounding Dino \cite{grounddino} to annotate masks with semantic information. 
Moreover, it is constrained to single-query segmentation for each mask, thereby limiting the ability to efficiently generate a whole segmentation map that necessitates rendering times on the order of minutes. 
Guo et al. \cite{guo2024semantic} projected 2D CLIP features onto 3D Gaussians via a spatial correspondence method that relied on rendered depth obtained from the depth rendering of 3D Gaussians. 
GSemSplat \cite{wang2024gsemsplat} tackled sparse-input 3D semantic understanding by utilizing ViT-based MASt3R \cite{mast3r} and an additional semantic MLP for feature prediction. Yet, its training pipeline involved O(N$^2$) pairwise point map computations, leading to high computational cost and limited scalability with increasing views. 
Jiao et al. \cite{jiao2024clip} proposed an image-text-3D Gaussian contrastive pretraining framework for 3D representation learning, but its performance depended on large-scale paired image-3D datasets.

Unlike previous approaches, our method addresses the inefficiency challenge by introducing a semantic attribute compactness strategy to compactly represent semantic Gaussian representations, facilitating fast training and inference. 
Additionally, we tackle the view-inconsistent issue through a novel 3D coherent regularization approach, enhancing the view consistency of semantics for coherent 3D semantic understanding.

\begin{figure*}[!t]
\centering
\includegraphics[width=\linewidth]{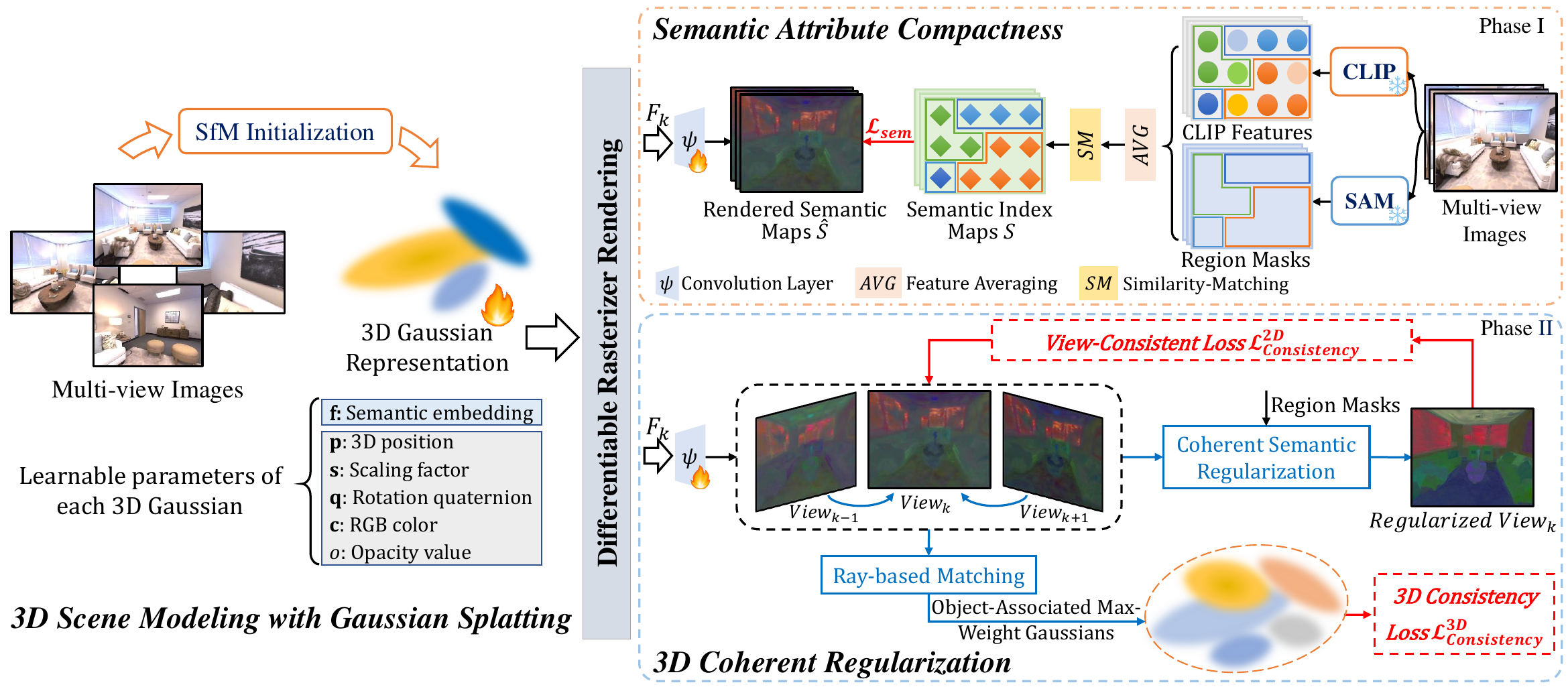}
\caption{
\textbf{Illustration of CLIP-GS optimization.} 
\textit{Left}: 
CLIP-GS represents the 3D scene with a collection of 3D Gaussians \cite{3dgs} with learnable attributes, specifically adding a \textit{semantic} attribute. 
\textit{Right}: 
First, multi-view images undergo feature extraction using the frozen CLIP model \cite{OpenAICLIP} and region mask generation with SAM \cite{sam}. 
We then optimize CLIP-GS in an end-to-end manner through two phases. 
In Phase I, we introduce \textbf{Semantic Attribute Compactness (SAC)} to capture the \textit{unified} semantics within each object, facilitating efficient optimization and rendering of semantic Gaussians. 
In Phase II, after training 3D Gaussians at certain iterations, we present \textbf{3D Coherent Regularization (3DCR)} to enhance 3D semantic consistency. 
3DCR leverages self-predicted semantics derived from CLIP-GS, refined by cross-view coherent regularization, to provide view-consistent supervision signals for optimizing Gaussians. 
Additionally, 3DCR identifies 3D Gaussian primitives associated with the same object through ray-based intersection matching and encourages their semantics to be similar. 
The color optimization process follows 3DGS \cite{3dgs} and is omitted for brevity. 
}
\label{fig:Overview} 
\end{figure*}

\section{Methodology}
\label{sec:methodology}

\subsection{Preliminary and Overview} \label{subsec:preliminary}
\textbf{Preliminary}. 
3D Gaussian Splatting (3DGS) \cite{3dgs} represents a 3D scene using a suite of 3D Gaussians, each parameterized by a 3D position $\mathbf{p} = \{x, y, z \} \in \mathbb{R}^3$, a 3D size scaling factor $\mathbf{s} \in \mathbb{R}^3$, a rotation quaternion $\mathbf{q} \in \mathbb{R}^4$, a color $\mathbf{c} \in \mathbb{R}^3$, and an opacity value $o \in \mathbb{R}$. 
All parameters are learnable and can be collectively symbolized by ${{\Theta}_{i}} = \{\mathbf{p}_{i}, \mathbf{s}_{i}, \mathbf{q}_{i}, \mathbf{c}_{i}, o_{i} \}$, where $i$ denotes the $i$-th Gaussian. 
To compute the pixel color ${C}$, 3DGS employs $\alpha$-blending point-based rendering by blending $\mathcal{N}$ Gaussians in the front-to-back depth order \cite{kopanas2021point}, formulated as:
\begin{equation}
{C(x_p)} = \sum_{i \in \mathcal{N}} T_i \alpha_i \mathbf{c}_i, 
\end{equation}
where $\mathcal{N}$ is the set of Gaussian primitives that overlap with the given pixel $x_p$. 
$\alpha_{i}$ is calculated by $\alpha_{i} = o_i G_i^{2D}(x_p)$, where $G_i^{2D}$ denotes the $i$-th Gaussian's 2D projection. 
The transmittance $T_i$ is defined as $T_i = \prod_{j=1}^{i-1} (1 - \alpha_j)$. 

To update the parameters of 3D Gaussians, Gaussian Splatting adopts a differentiable rendering technique that projects the Gaussians onto the 2D image plane \cite{ewaVolumeSplatting} and optimizes them using color supervision. 
The reconstruction loss is defined as the discrepancy between the rendered image $\hat{I}$ and the ground truth image $I$: 
\begin{equation}
\mathcal{L}_{rgb} = (1-\lambda)\mathcal{L}_{1}(\hat{I}, I) + \lambda\mathcal{L}_{D-SSIM}(\hat{I}, I), 
\end{equation}
where $\lambda$ is set to 0.2 \cite{3dgs}. 
3DGS has shown its effectiveness in 3D scene reconstruction. 
In this work, we propose \textbf{CLIP-GS}, a novel approach that extends 3DGS toward semantic understanding of complex 3D indoor scenes.

\textbf{Overview}.  
The overall framework of our CLIP-GS is illustrated in Fig. \ref{fig:Overview}. 
Given a set of $N$ posed images $I = \{ I_i \}_{i=1}^{N} \in \mathbb{R}^{H \times W \times 3} $, CLIP-GS represents the 3D scene using 3D Gaussian primitives with learnable attributes, specifically adding a semantic attribute. 
To enable 3D semantic understanding, we optimize 3D Gaussians guided by multi-view CLIP features through two key components: \textbf{Semantic Attribute Compactness (SAC)} and \textbf{3D Coherent Regularization (3DCR)}. 
SAC constructs compact semantic Gaussian representations to support efficient semantic understanding (Section \ref{subsec:sac}), while 3DCR introduces 2D and 3D-level semantic-consistent constraints to enhance coherent segmentation results (Section \ref{subsec:3DCR}). 
We elaborate on each component in the following sections.

\subsection{Semantic Attribute Compactness} \label{subsec:sac}
To render novel views with semantic information via the Gaussian Splatting rendering, 
an intuitive method is attaching a learnable semantic parameter $\mathbf{f}_{i} \in \mathbb{R}^D$ to each Gaussian, and apply $\alpha$-blending to compute the pixel-wise rendered feature ${F}$. 
This process can be formulated as: 
\begin{equation} 
{F(x_p)} = \sum_{i \in \mathcal{N}} T_i \alpha_i \mathbf{f}_i \in \mathbb{R}^D, \label{eq_Semalpha}
\end{equation}
where $D$ denotes the dimension of rendered features, typically set to a high value, such as 512, to align the CLIP semantic feature dimension for optimizing $\mathbf{f}_{i}$. 
However, this direct embedding of high-dimensional parameters into millions of Gaussians for semantic modeling, such as Feature 3DGS \cite{feature3dgs}, significantly decreases rasterization efficiency in Gaussian Splatting, leading to constrained rendering performance.

To tackle this efficiency challenge, we introduce \textbf{Semantic Attribute Compactness (SAC)}. 
The key insight of SAC is leveraging the inherently \textit{unified} semantic meaning of the identical object for efficient representation. 
Concretely, SAC represents each object using a single representative CLIP feature, and further employs low-dimensional semantic indices to encapsulate the high-dimensional CLIP features. 
By leveraging low-dimensional indices as supervision, we thus attach low-dimensional semantic embeddings to 3D Gaussians for rasterization, resulting in a more efficient semantic rendering process.

Specifically, for $k$-th training view $I_k$, we transform the CLIP feature $\bar{F}_k$ $\in \mathbb{R}^{D \times H \times W}$ into a representative version $\hat{F}_k$ $\in \mathbb{R}^{D \times M}$, where $M$ denotes the number of objects in $I_k$. 
To achieve this, we harness the powerful Segment Anything Model (SAM) \cite{sam} to yield region masks ${R}_k = \{ {R}_k^{q} \}_{q=1}^{M}$ over the image $I_k$. 
For each region, we compute the weighted average of the CLIP feature in the spatial dimension, treating it as the unified feature to represent the semantics of this region uniformly. 
This process yields the representative CLIP feature $\hat{F}_k$ $\in \mathbb{R}^{D \times M}$, where $M$ typically ranges from a few dozen to a hundred, significantly smaller than the image size $H \times W$. 
Since the features within each region are still high-dimensional yet semantically consistent, we utilize a low-dimensional semantic index to efficiently represent the unified semantic feature of each region. 
To assign these indices for each region, we adopt a similarity-matching method that computes the cosine similarity between representative CLIP features $\hat{F}_k$ and a set of text features $T$, producing the semantic index map $S_k$ as: 
\begin{equation}
S_k = \mathbf{argmax} (\mathbf{cos} (\hat{F}_k, T)),  
\end{equation} 
where $T$ is obtained by encoding a set of text descriptions using the CLIP text encoder. 
In this way, each position within a region of the semantic index map $S_k \in \mathbb{R}^{1 \times H \times W}$ shares a consistent, low-dimensional semantic index. 
This method ensures a reliable and robust matching relationship in the CLIP feature space, generating effective semantic index maps for 3D Gaussian optimization and CLIP feature retrieval.

Leveraging SAC, we can transform semantic Gaussian representation learning into a low-dimensional space. Thus, we embed a low-dimensional, learnable semantic parameter $\mathbf{f}_{i} \in \mathbb{R}^d$ into each 3D Gaussian to efficiently construct the 3D semantic field. 
Specifically, we adopt the $\alpha$-blending rendering pipeline to project 3D Gaussians onto the 2D image plane and obtain the rendered feature map $F \in \mathbb{R}^{d \times H \times W}$. Each pixel-wise feature is computed as $F(x_p) = \sum_{i \in \mathcal{N}} T_i \alpha_i \mathbf{f}_i $, where $\mathcal{N}$ denotes the set of Gaussians contributing to pixel location $x_p$. 
Then, we use a trainable, lightweight convolution layer $\psi$ to produce the rendered semantic map $\hat{S}$, which can be formulated as: 
\begin{equation}
\hat{S} = \psi(F).  
\end{equation} 
The rendered semantic map $\hat{S}$ is supervised using the semantic index map $S$, optimizing the learnable semantic embedding per Gaussian, and the semantic loss is defined as: 
\begin{equation}
\mathcal{L}_{sem} =  \mathcal{L}_{ce} (\hat{S}, {S} ).
\end{equation} 
By enhancing 3D Gaussians with efficient semantic modeling, our approach enables efficient rendering while exhibiting effective segmentation results.

\begin{figure*}[!t]
\centering
\includegraphics[width=.98\linewidth]{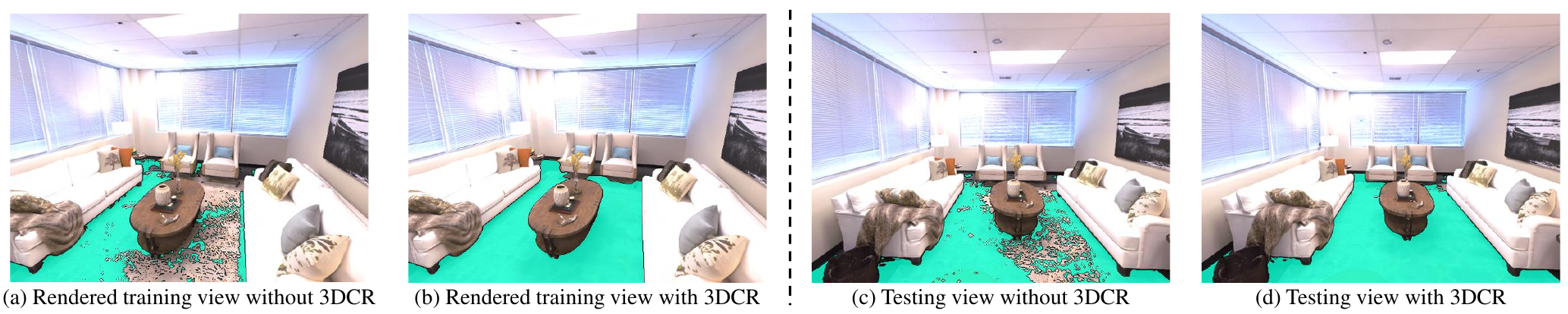}
\caption{
Illustration of rendered segmentation maps with the text query "Rug". 
(a) and (b) correspond to training views, while (c) and (d) pertain to testing views. 
In (a), the rendered result appears ambiguous when immutably employing 2D CLIP semantics for Gaussian optimization. 
Conversely, (b) shows that leveraging 3DCR can provide coherent semantic constraints to supervise Gaussians, leading to more precise results (d).  
For more details, refer to Sec. \ref{subsec:3DCR}. 
}
\label{fig:Vis_3DCR}
\end{figure*}

\subsection{3D Coherent Regularization} \label{subsec:3DCR}
Having achieved efficient semantic representations through semantic attribute compactness, however, immutably employing view-inconsistent CLIP semantics to optimize Gaussians leads to ambiguous and subpar rendering results, as illustrated in (a) and (c) of Fig. \ref{fig:Vis_3DCR}. 
This limitation stems from the 2D CLIP model's inherent difficulty in maintaining consistent object identities across different views, inadequately enforcing multi-view semantic consistency constraints. 
To tackle this limitation, we propose a \textbf{3D Coherent Regularization (3DCR)} approach to enhance semantic-consistent constraints at 2D and 3D levels.

\begin{figure}[!t]
\centering
\includegraphics[width=.98\linewidth]{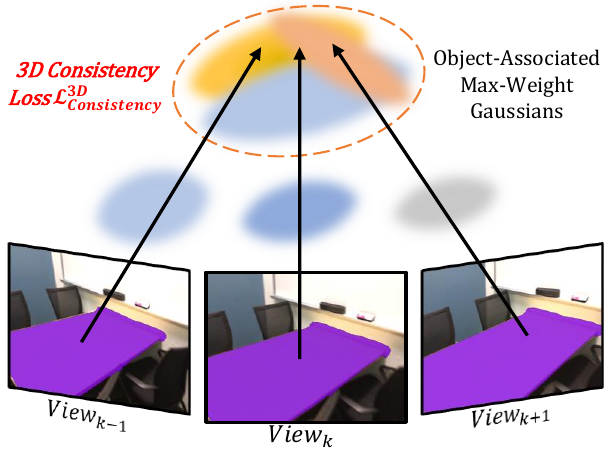}
\caption{
Illustration of the consistency regularization for 3D Gaussians in 3D Coherent Regularization (3DCR). 
}
\label{fig:Vis_3DCR_matching}
\end{figure}

\textbf{Consistency Regulrization for 2D Views.}
Inspired by the inherently consistent semantics rendered from trained 3D models, we leverage enhanced self-predicted semantics derived from trained 3D Gaussians as view-consistent supervision signals for imposing cross-view coherent semantic constraints. 

Specifically, after several training iterations, we render a semantic map $\hat{S}_k$ from the current view $I_k$. We then introduce a \textit{coherent semantic regularization} by incorporating semantic cues from adjacent training views, to eliminate semantic ambiguity and establish view-consistent supervision signals for the current view $I_k$. 

To achieve this, we treat the sequence of multi-view training images as temporally adjacent frames, akin to a video, and use a zero-shot tracker \cite{deva} to associate SAM masks across adjacent views. 
Next, we render semantic maps $\hat{S}_{k-1}$ and $\hat{S}_{k+1}$ at the preceding and subsequent views, respectively.
For the $q$-th region ${R}_k^{q}$ in the $k$-th view, we aggregate semantic information from corresponding regions ($R_{k-1}^{q}, R_{k+1}^{q}$) from adjacent views, and perform coherent regularization to unify the $\hat{S}_k^q$ within the region $R_k^{q}$. 
This coherent regularization is achieved through a majority voting strategy. 
Intuitively, this strategy unifies the region-based semantic representation by selecting the most frequently occurring semantic index across views, thereby forming a consistent semantic representation ${Z}_k^q$, which can be formulated as: 
\begin{equation}
{Z}_{k}^{q} = \underset{c}{\mathbf{argmax}} \sum_{v \in { U_k^{q}} } \mathbbm{1} \{ \mathbf{argmax}(\psi({\hat{S}_{k}^{v}})) = c \}, 
\end{equation} 
where $ U_k^{q} \equiv \{ R_{k-1}^{q}, R_k^{q}, R_{k+1}^{q} \} $ denotes the set of corresponding region masks across the current and adjacent views, providing enhanced coherent semantic regularization. 
$c$ indicates the $c$-th category. 
The indicator function $\mathbbm{1} \{ \cdot \}$ returns 1 if the condition matches category $c$ at a given pixel, and 0 otherwise. 
Similarly, this coherent regularization is applied to all regions in the rendered semantic map $\hat{S}_{k}$. 
As shown in (b) of Fig. \ref{fig:Vis_3DCR}, this regularization effectively improves the semantic consistency, such as the \textit{rug} area in the current view.
Consequently, we utilize this view-consistent supervision signal ${Z}$ to train 3D Gaussians, rather than vanilla 2D CLIP semantics: 
\begin{equation}
\mathcal{L}_{consistency}^{2D} = \mathcal{L}_{ce} (\hat{S}, {Z} ). 
\label{eq_2d_consistency}
\end{equation}
Moreover, in each iteration, this coherent semantic regularization is applied to a single training view, ensuring training efficiency. 
With progressively enhanced view-consistent constraints, the semantic consistency of 3D Gaussians improves over time, which, in turn, reinforces the regularization in subsequent iterations.
As a result, this iterative refinement facilitates more coherent segmentation results, as evidenced in the experiments.

\textbf{Consistency Regularization for 3D Gaussians.}
Beyond 2D constraints, we further introduce a feature similarity constraint among 3D Gaussian primitives associated with the same object. 
Since the rendering results are predominantly determined by maximally-weighted Gaussian primitives (as formulated in Eq. \eqref{eq_Semalpha}), we encourage semantic similarity 
among these dominant Gaussians associated with the same object. 

For instance, as shown in Fig. \ref{fig:Vis_3DCR_matching}, given an object region ${R}_k^{q}$ in $k$-th view $I_k$, we first gather pixel locations from the corresponding object masks $\{ R_{k-1}^{q}, R_k^{q}, R_{k+1}^{q} \} $ across the current and adjacent views. 
For each of these pixels, we identify the maximally contributing Gaussian along the ray cast into 3D space, yielding a matched set of Gaussian primitives $\mathbf{G}$ = $\{{\Theta}_{i} \}_{i=1}^n$ that intersect rays from all three views. 
These matching, maximally weighted Gaussians associated with the same object are then encouraged to have similar semantic representations. 
To implement this, we minimize the distance between their features $\mathbf{H} = \{ \mathbf{f}_i \in \mathbf{G} \}_{i=1}^n$ and their corresponding cluster feature $\mathbf{M}$, computed via mean-pooling over set $\mathbf{G}$. 
The 3D feature similarity constraint is enforced through a KL-divergence-based loss as follows: 
\begin{equation}
\mathcal{L}_{consistency}^{3D} = \mathcal{L}_{kl}( \mathbf{M} || \mathbf{H} ) = \sum_{i = 1}^{n} \mathbf{M} \mathrm{log}( \frac{\mathbf{M}}{ \mathbf{f}_i } ). 
\label{eq_3d_consistency}
\end{equation}

This loss encourages 3D Gaussians associated with the same object to share similar semantic features by minimizing their divergence from a pooled semantic anchor, thereby enhancing 3D semantic consistency. 
In summary, the overall 3D coherent regularization loss contains 2D and 3D-level constraints, formulated as:
\begin{equation}
\mathcal{L}_{consistency} = \mathcal{L}_{consistency}^{2D} + \mathcal{L}_{consistency}^{3D}. 
\end{equation}

\subsection{Overall Training with Progressive Densification Regulation Scheme} \label{subsec:training} 
Our overall training scheme serves two objectives. 
First, to establish a semantic Gaussian radiance field for accurate and view-consistent semantic understanding of 3D scenes.  
Second, to regulate the quantity of Gaussian primitives for more efficient semantic Gaussian radiance field acquisition.

Specifically, the entire framework is trained end-to-end using reconstruction loss $\mathcal{L}_{rgb}$ and semantic loss $\mathcal{L}_{s}$ across two phases. 
In phase I, the semantic embeddings of 3D Gaussians are optimized utilizing $\mathcal{L}_{sem}$. 
In phase II, after training the 3D Gaussians a few iterations $\mathcal{T}$, the 3D coherent regularization loss $\mathcal{L}_{consistency}$ replaces the $\mathcal{L}_{sem}$ for semantic consistency enhancement. 
The overall training process can be formulated as: 
\begin{align}
\mathcal{L} = 
\left\{ 
    \begin{array}{lc}
        \mathcal{L}_{rgb} + \mathcal{L}_{sem} & \mathbf{Iteration} \leq \mathcal{T}
        \\
        \mathcal{L}_{rgb} + \mathcal{L}_{consistency} & \mathbf{Iteration} >  \mathcal{T}. \\
    \end{array}
\right.
\end{align}

In addition to achieving precise 3D semantic understanding, we introduce a \textbf{Progressive Densification Regulation (PDR)} strategy to regulate the amount of Gaussians, aiming to improve efficiency while upholding high-quality scene representations. 
Specifically, in vanilla 3DGS, loss computation initially operates at full image resolution with fixed densification parameters (e.g., threshold and interval) for 3D Gaussians. This fixed training scheme often results in a proliferation of Gaussian primitives at the early training process, as the densification strategy outlined in \cite{3dgs} rapidly causes new Gaussians to further split or clone before adequate optimization, leading to redundant Gaussian generation and compromising rendering efficiency.

To mitigate primitive proliferation, when employing our PDR, Gaussians undergo initial optimization at lower image resolutions with reduced densification frequency and elevated densification threshold. 
As training progresses, the image resolution and densification frequency gradually increase while the densification threshold decreases, ultimately reaching full resolution and default densification parameters. 
Experimental results show that PDR not only effectively prevents excessive proliferation of Gaussians in the early training stage to enhance rendering efficiency, but also constructs a more effective Gaussian radiance field.

\section{Experiment} \label{sec:experiment}

\subsection{Experiment Setup} 
\textbf{Evaluation Datasets.} 
We evaluate our approach on three multi-view indoor scene datasets that are extensively used in 3D scene reconstruction and segmentation, including 3DOVS \cite{3DOVS}, ScanNet \cite{scannet}, and Replica \cite{replica}. 
\begin{itemize}
    \item 3DOVS is a real-world dataset containing diverse objects captured in various poses and backgrounds. The scenes featured in 3DOVS consist of 28 to 37 images with a face-forward orientation, whose views are sampled in an "outside-in" manner, resulting in notable overlap across views. We conduct experiments on four scenes (Bed, Sofa, Lawn, and Bench) for evaluation. 
    We follow the experimental protocol outlined in 3DOVS \cite{3DOVS}, using posed training images for optimizing semantic fields and semantic labels from the testing set for evaluation. 
    \item ScanNet is a real-world dataset that offers a variety of indoor scenes and provides camera poses obtained via BundleFusion \cite{Bundlefusion} and semantic segmentation labels. The experiments involve four scenes (Scene0004, Scene0389, Scene0494, and Scene0693), each with 233 to 289 images captured along predefined trajectories. Every 10-$th$ is reserved for evaluation, with the remaining images used for training.  
    \item Replica is a synthetic dataset that includes high-fidelity indoor scenes with photorealistic textures and per-primitive semantic classes. 
     We select six scenes (Room0, Room1, Room2, Office0, Office2, and Office4) for evaluation. For each scene, images are captured along a defined trajectory, with every 10-$th$ image designated for evaluation and the rest used for training.   
\end{itemize}

To further evaluate the robustness of our method, we introduce a \textit{sparse-view} benchmark, specifically targeting challenging "inside-out" indoor scenarios in ScanNet and Replica. For each scene, we evenly sample 30 images. Of these, every 10-$th$ image is designated as a test image, and the remaining images serve as training data.

\noindent
\textbf{Evaluation Metrics.} 
We evaluate segmentation performance on novel views using two metrics: mean Intersection over Union (mIoU) and mean Pixel Accuracy (mAcc). 
Moreover, to assess the quality of novel view images, we leverage Peak Signal-to-Noise Ratio (PSNR), Structural Similarity Index Measure (SSIM), and Learned Perceptual Image Patch Similarity (LPIPS) \cite{lpips} to assess the quality of novel view images. 


\begin{table*}[t]
\renewcommand{\arraystretch}{1.2}
\caption{Comparison with state-of-the-art methods on segmentation results of novel views across different scenes from the ScanNet dataset \cite{scannet}, under multi-view training data conditions. 
Our proposed approach demonstrates superior performance. 
} 
\label{table:scene_multieview_scannet}
\centering 
\begin{adjustbox} {width=.9\linewidth}
\begin{tabular}{ r | cc | cc | cc | cc | cc }
\Xhline{3\arrayrulewidth}
\multirow{2}{*}{Method} & \multicolumn{2}{c|}{Scene0004} & \multicolumn{2}{c|}{Scene0389} & \multicolumn{2}{c|}{Scene0494} & \multicolumn{2}{c|}{Scene0693} & \multicolumn{2}{c}{Average} \\ 
& mIoU $\uparrow$ & mAcc $\uparrow$ & mIoU $\uparrow$ & mAcc $\uparrow$ & mIoU $\uparrow$ & mAcc $\uparrow$ & mIoU $\uparrow$ & mAcc $\uparrow$ & mIoU $\uparrow$ & mAcc $\uparrow$  \\  
\midrule[0.5pt]
LERF   \cite{LERF}    & 7.137 & 12.139  & 20.102 & 62.025 & 17.624 & 35.079 & 16.531 & 51.933 & 15.349 & 40.294  \\ 
3DOVS  \cite{3DOVS}   & 9.763  & 15.666 & 20.596 & 64.099 & 18.556 & 36.223 & 22.292 & 54.141 & 17.802 & 42.532  \\ 

Feature 3DGS \cite{feature3dgs}  & 20.907 & 52.557  & 19.978 & 72.902 & 26.204 & 52.191 & 12.236 & 39.614 & 19.831  & 54.316 \\ 
LangSplat  \cite{langsplat}      & 30.318 & 69.623  & 18.936 & 68.003 & 21.167 & 47.796 & 16.526 & 33.789 & 21.737  & 54.803 \\ 
\midrule[0.5pt]
Ours    
& \textbf{40.123} & \textbf{78.851} & \textbf{39.761} & \textbf{89.393} 
& \textbf{58.543} & \textbf{88.224} & \textbf{33.303} & \textbf{62.674} 
& \textbf{42.932} & \textbf{79.786} \\ 
\Xhline{3\arrayrulewidth}
\end{tabular}
\end{adjustbox}
\end{table*}

\begin{table*}[t]
\renewcommand{\arraystretch}{1.2}
\caption{Comparison with state-of-the-art methods on segmentation results of novel views across different scenes from the Replica dataset \cite{replica}, under multi-view training data conditions. 
Our proposed approach demonstrates superior performance. 
} 
\label{table:scene_multieview_replica}
\centering 
\begin{adjustbox} {width=\linewidth}
\begin{tabular}{ r | cc | cc | cc | cc | cc | cc | cc }
\Xhline{3\arrayrulewidth}
\multirow{2}{*}{Method} & \multicolumn{2}{c|}{Room0} & \multicolumn{2}{c|}{Room1} & \multicolumn{2}{c|}{Room2} & \multicolumn{2}{c|}{Office0} & \multicolumn{2}{c|}{Office2} & \multicolumn{2}{c|}{Office4} & \multicolumn{2}{c}{Average} \\ 
& mIoU $\uparrow$ & mAcc $\uparrow$ & mIoU $\uparrow$ & mAcc $\uparrow$ & mIoU $\uparrow$ & mAcc $\uparrow$ & mIoU $\uparrow$ & mAcc $\uparrow$ & mIoU $\uparrow$ & mAcc $\uparrow$ & mIoU $\uparrow$ & mAcc $\uparrow$ & mIoU $\uparrow$ & mAcc $\uparrow$ \\  
\midrule[0.5pt]
LERF   \cite{LERF}    & 7.615 & 35.346 & 14.043 & 30.218 & 6.552 & 22.468 & 3.369  & 4.887 & 7.077 & 16.884 & 11.056 & 22.948 & 8.285 & 22.125 \\
3DOVS  \cite{3DOVS}   & 7.733 & 36.667 & 16.016 & 35.957 & 7.547 & 25.675 & 3.843  & 6.487 & 7.070 & 18.409 & 12.276 & 20.435 & 9.081 & 23.938 \\ 
Feature 3DGS \cite{feature3dgs}  & 8.794  & 39.916 & 10.226 & 32.024 & 11.909 & 40.558 & 8.120  & 21.588 & 11.218 & 41.776 & 13.537 & 43.260 & 10.634 & 36.520 \\ 
LangSplat  \cite{langsplat}      & 11.932 & 45.703 & 15.706 & 47.586 & 19.658 & 72.577 & 9.968 & 36.708 & 17.805 & 55.518 & 16.445 & 51.356 & 15.252 & 51.575 \\ 
\midrule[0.5pt]
Ours 
& \textbf{27.249} & \textbf{74.129} & \textbf{28.358} & \textbf{72.140} 
& \textbf{23.082} & \textbf{76.659} & \textbf{20.932} & \textbf{38.489} 
& \textbf{40.312} & \textbf{90.452} & \textbf{29.882} & \textbf{56.991} 
& \textbf{28.302} & \textbf{68.143} \\ 
\Xhline{3\arrayrulewidth}
\end{tabular}
\end{adjustbox}
\end{table*}

\subsection{Implementation Details}  
\textbf{Data Pre-processing.}
For CLIP feature extraction, we utilize the CLIP ViT-B/16 model \cite{OpenAICLIP} as the base experimental setting. Moreover, we provide CLIP-LSeg ViT-L/16 \cite{LSeg} that is used in Feature 3DGS \cite{feature3dgs}, for more comprehensive comparisons.  
To acquire the SAM's masks, we employ the SAM ViT-H model \cite{sam}. 
Specifically, we deploy SAM to automatically generate masks for each training image. Point prompts are sampled on a 32 × 32 grid, with a minimum mask region area set to 100 and a bounding box IoU threshold of 0.7.
The CLIP features and SAM's masks mentioned above are pre-computed offline.

\noindent
\textbf{Training.}
We implement our approach using PyTorch and Gaussian Splatting \cite{3dgs} on an NVIDIA A100 GPU, adhering to the default parameter settings in Gaussian Splatting for scene reconstruction. 
To enable scene understanding, we augment each 3D Gaussian with learnable semantic parameters of dimension $d=3$ and modify the CUDA kernel to incorporate semantic rasterization while maintaining reconstruction quality. 

During training, the learning rates for the semantic parameters and convolution layer are set to 2.5$e^{-3}$ and 5$e^{-4}$, respectively. 
Adam optimizer is used to train our model for 30k iterations, with each scene taking approximately 20 minutes to train. 
The similarity-matching computation in SAC is conducted offline to generate semantic index maps before training. 
The hyperparameter $\mathcal{T}$ is set to 15k to leverage the 3D coherent regularization method. 
For PDR, we start with a training image resolution scale factor of 0.5, a densification interval of 200 iterations, and a densification threshold scale factor of 1.5. 
We then gradually increase the resolution to 1.0, while the densification interval and threshold are adjusted toward their default values following a cosine scheduling scheme. 
The resolution increase occurs during the first 7k iterations, and the densification adjustment takes place within the initial 4k iterations.


\begin{figure*}[!t]
\centering
\includegraphics[width=.99\linewidth]{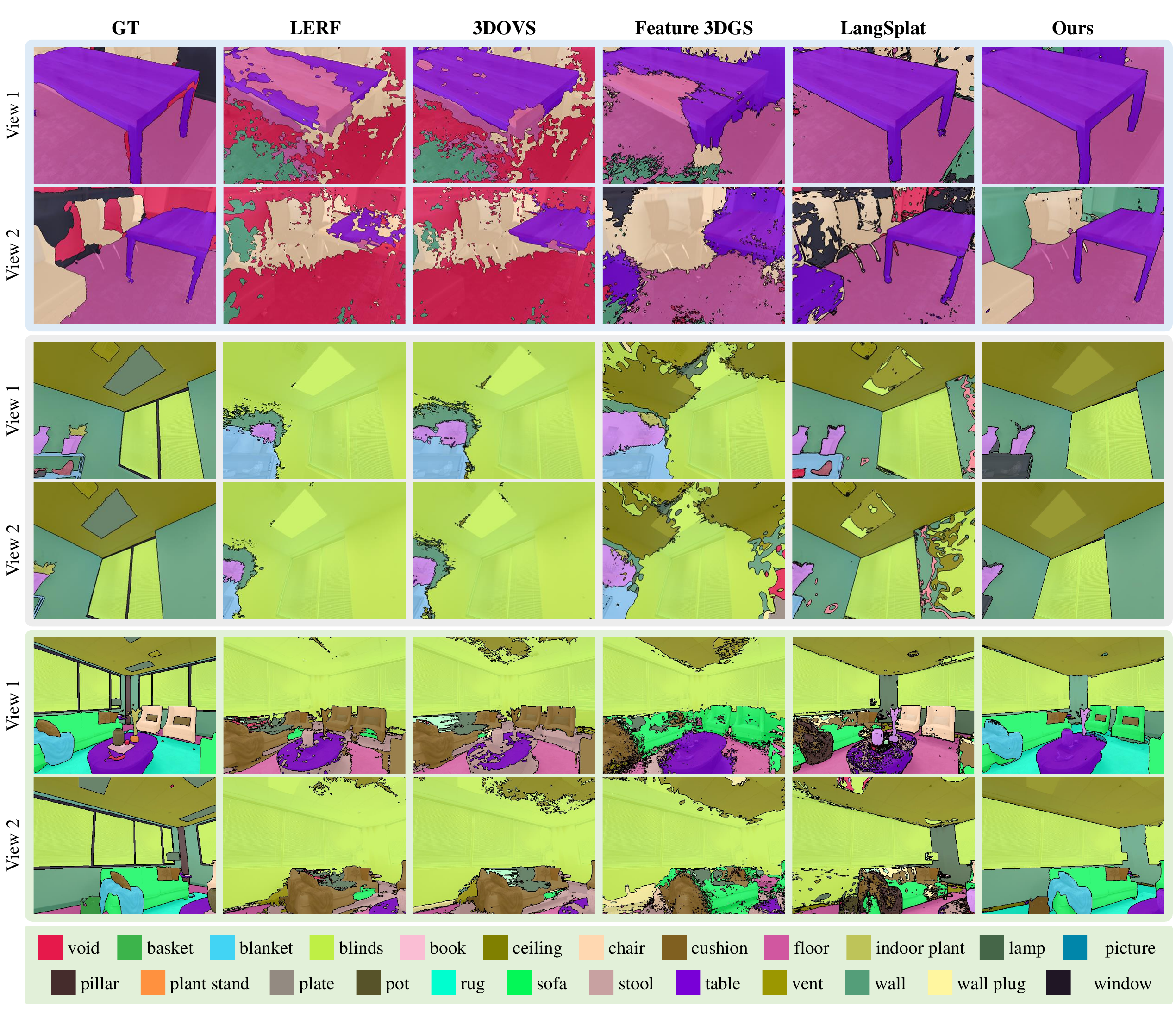}
\caption{
Visual segmentation results of novel view under multi-view training data conditions. While current methods produce ambiguous segmentation results due to the absence of 3D consistency constraints, our method achieves more precise and view-consistent results across various views. 
}
\label{fig:Vis_multiview_seg}
\end{figure*}

\subsection{Comparison with State-of-the-art Methods} \label{section:comparison_SOTA}
\subsubsection{\textbf{Results on the ScanNet Dataset}}
We compare our approach with NeRF-based methods: LERF \cite{LERF} and 3DOVS \cite{3DOVS}, and 3DGS-based methods: Feature 3DGS \cite{3dgs} and LangSplat \cite{langsplat}.  

We present the quantitative results in Table \ref{table:scene_multieview_scannet}. 
Our method consistently outperforms other competitors in segmentation accuracy across diverse datasets. 
Notably, our approach exhibits a significant improvement in mIoU of 21.20\% over the second-best method. 
These results indicate that our method effectively leverages semantic-consistent constraints for 3D Gaussian optimization, achieving precise 3D semantic understanding. 

The qualitative analysis, as presented in the first two rows of Fig. \ref{fig:Vis_multiview_seg}, demonstrates that most current methods struggle to accurately segment object boundaries. While LangSplat employs SAM-generated region masks to enhance the object boundaries (e.g., \textit{Table}), it easily yields ambiguous and noisy segmentation results due to the lack of view-consistent semantic constraints.  
In contrast, our approach exhibits spatially continuous and view-consistent results, stemming from our 3D coherent regularization approach, which effectively incorporates view-consistent semantic constraints to ensure coherent and accurate results.

\subsubsection{\textbf{Results on the Replica Dataset}}
We further evaluate the open-vocabulary segmentation performance through testing on synthetic scenes from Replica.  

Table \ref{table:scene_multieview_replica} reports the quantitative results on the Replica datasets. It can be seen that our method also achieves superior performance in segmentation accuracy across various scenes.  
Specifically, our approach achieves remarkable mIoU improvements of 13.05\% compared to the second-best method. 
These findings demonstrate the effectiveness of our method for accurate 3D semantic understanding. 

As shown in the last four rows of Fig. \ref{fig:Vis_multiview_seg}, we present qualitative results produced by our method and other approaches.  
It can be seen that existing methods face challenges in delivering precise and complete segmentation results (e.g., \textit{Blinds}), our method consistently produces accurate and coherent segmentation results. 
These improvements stem from our 3D coherent regularization approach, which effectively enforces 3D consistency constraints to improve cross-view semantic consistency.

\begin{table*}[t] 
\renewcommand{\arraystretch}{1.1}
\caption{
Comparison with state-of-the-art methods on segmentation results of novel views across various scenes from the 3DOVS dataset \cite{3DOVS}. 
Our proposed approach demonstrates superior open-world segmentation performance. 
}
\label{table:SOTA_3DOVS}
\centering
\begin{adjustbox} {width=.9\linewidth}
\begin{tabular}{ r | c c | c c | c c | c c | c c }
\Xhline{3\arrayrulewidth}
\multirow{2}{*}{Method} 
& \multicolumn{2}{c|}{Bed}  & \multicolumn{2}{c|}{Sofa} & \multicolumn{2}{c|}{Lawn} & \multicolumn{2}{c|}{Bench} & \multicolumn{2}{c}{Average}  \\
& mIoU $\uparrow$ & mAcc $\uparrow$ & mIoU $\uparrow$ & mAcc $\uparrow$ & mIoU $\uparrow$ & mAcc $\uparrow$ & mIoU $\uparrow$ & mAcc $\uparrow$ & mIoU $\uparrow$ & mAcc $\uparrow$ \\  
\midrule[0.5pt]
FFD    \cite{DFF}     & 56.6   & 86.9  & 3.7   & 9.5    & 42.9   & 82.6   & 6.1   & 42.8  & 27.3   & 55.5 \\  
LERF   \cite{LERF}    & 73.5 & 86.9 & 27.0 & 43.8 & 73.7 & 93.5 & 53.2 & 79.7  & 56.9 & 76.0 \\ 
3DOVS   \cite{3DOVS}  & 89.5 & 96.7 & 74.0 & 91.6 & 88.2 & 97.3 & 89.3 & 96.3  & 85.3 & 95.5 \\ 
LangSplat   \cite{langsplat}  & 92.5 & 99.2 & 90.0 & 97.9 & 96.1 & \textbf{99.4} & 94.2 & 98.6 & 93.2 & 98.8 \\ 
\midrule[0.5pt] 
Ours  & \textbf{97.2} & \textbf{99.3} & \textbf{94.1} & \textbf{98.4} & \textbf{96.5} & \textbf{99.4} & \textbf{94.8} & \textbf{98.7} & \textbf{95.6} & \textbf{99.0} \\ 
\Xhline{3\arrayrulewidth}
\end{tabular}
\end{adjustbox}
\end{table*}

\begin{figure*}[!t]
\centering
\includegraphics[width=.99\linewidth]{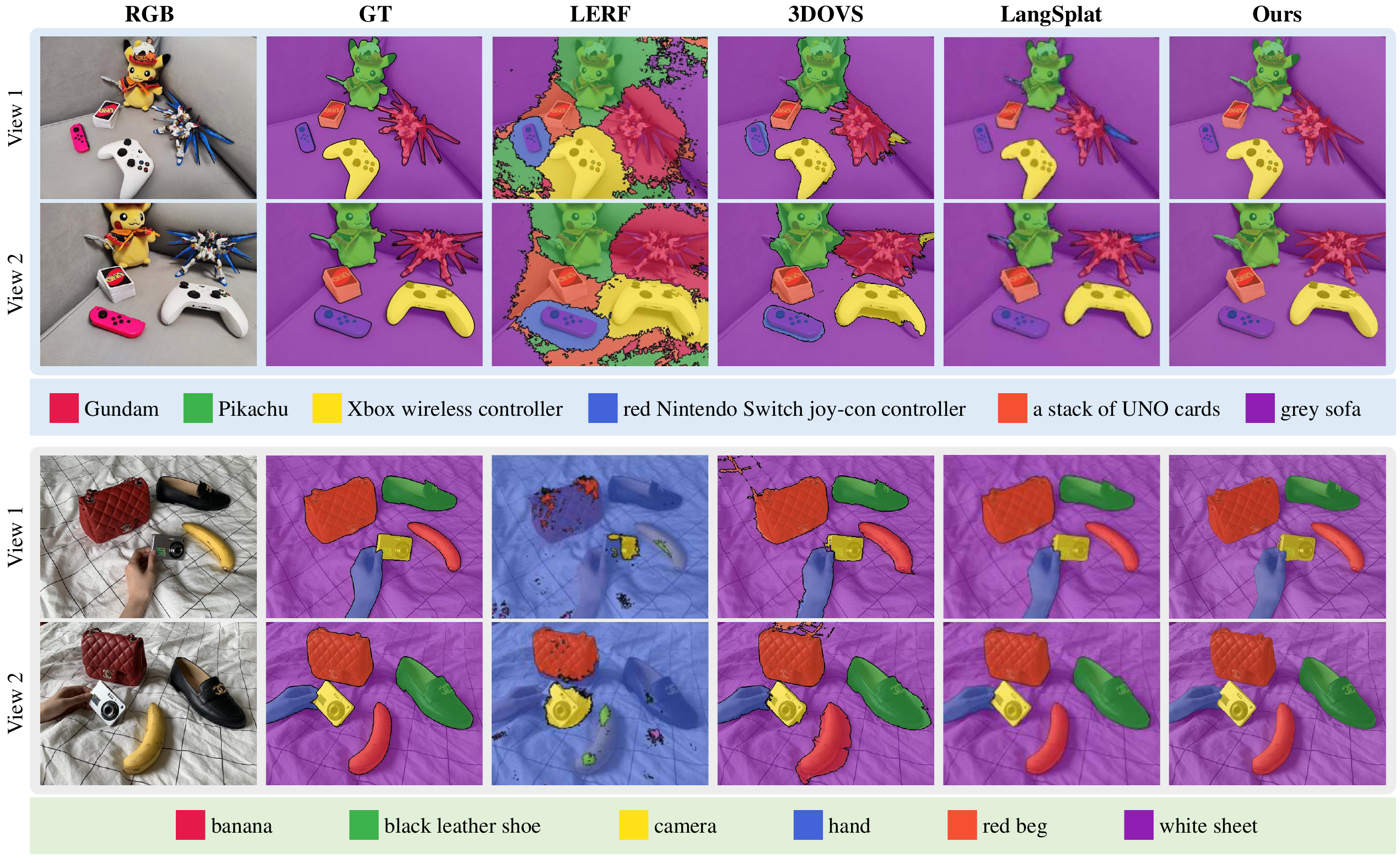}
\caption{
Visual segmentation results on various scenes from the 3DOVS dataset \cite{3DOVS}. 
Our approach maintains accurate and stable generalization performance in face-forwarding scenes. 
}
\label{fig:Vis_3dovs_seg}
\end{figure*}

\subsubsection{\textbf{Results on the 3DOVS Dataset}}
Additionally, we evaluate our method on diverse face-forwarding scenes from the 3DOVS dataset. 
Quantitative comparisons presented in Table \ref{table:SOTA_3DOVS}, coupled with qualitative visualizations in Fig. \ref{fig:Vis_3dovs_seg}, demonstrate that our approach achieves superior performance, obtaining precise and finer boundaries for each object. 
These results substantiate the stable generalization performance of our approach when dealing with face-forwarding scenes.

\begin{table*}[t]
\renewcommand{\arraystretch}{1.2}
\caption{Comparison with state-of-the-art methods on segmentation results of novel views across different scenes from the ScanNet dataset \cite{scannet}, under sparse-view training data conditions. 
Our proposed approach maintains results. 
} 
\label{table:scene_sparseview_scannet}
\centering 
\begin{adjustbox} {width=.9\linewidth}
\begin{tabular}{ r | cc | cc | cc | cc | cc }
\Xhline{3\arrayrulewidth}
\multirow{2}{*}{Method} & \multicolumn{2}{c|}{Scene0004} & \multicolumn{2}{c|}{Scene0389} & \multicolumn{2}{c|}{Scene0494} & \multicolumn{2}{c|}{Scene0693} & \multicolumn{2}{c}{Average} \\ 
& mIoU $\uparrow$ & mAcc $\uparrow$ & mIoU $\uparrow$ & mAcc $\uparrow$ & mIoU $\uparrow$ & mAcc $\uparrow$ & mIoU $\uparrow$ & mAcc $\uparrow$ & mIoU $\uparrow$ & mAcc $\uparrow$  \\  
\midrule[0.5pt]
LERF   \cite{LERF}    & 8.838  & 9.259  & 15.528 & 74.926 & 20.594 & 30.911 & 11.276 & 39.841 & 14.059 & 38.734 \\ 
3DOVS  \cite{3DOVS}   & 12.726 & 30.569 & 12.754 & 62.287 & 14.989 & 24.847 & 16.438 & 44.634 & 14.227 & 40.584 \\  
Feature 3DGS \cite{feature3dgs}  & 22.414 & 51.851 & 19.341 & 66.565 & 23.914 & 46.779 & 10.742  & 30.438 & 19.103 & 48.908 \\  
LangSplat  \cite{langsplat}      & 16.448 & 42.199 & 7.724 & 32.094 & 18.962 & 44.276  & 7.443 & 20.192 & 12.644 & 34.690  \\   
\midrule[0.5pt]
Ours 
& \textbf{25.849} & \textbf{53.013} & \textbf{41.507} & \textbf{84.122} 
& \textbf{59.041} & \textbf{85.188} & \textbf{44.656} & \textbf{71.825} 
& \textbf{42.763} & \textbf{73.537} \\ 
\Xhline{3\arrayrulewidth}
\end{tabular}
\end{adjustbox}
\end{table*}

\begin{table*}[t]
\renewcommand{\arraystretch}{1.2}
\caption{Comparison with state-of-the-art methods on segmentation results of novel views across different scenes from the Replica dataset \cite{replica}, under sparse-view training data conditions. 
Our proposed approach maintains superior results. 
} 
\label{table:scene_sparseview_replica}
\centering 
\begin{adjustbox} {width=\linewidth}
\begin{tabular}{ r | cc | cc | cc | cc | cc | cc | cc }
\Xhline{3\arrayrulewidth}
\multirow{2}{*}{Method} & \multicolumn{2}{c|}{Room0} & \multicolumn{2}{c|}{Room1} & \multicolumn{2}{c|}{Room2} & \multicolumn{2}{c|}{Office0} & \multicolumn{2}{c|}{Office2} & \multicolumn{2}{c|}{Office4} & \multicolumn{2}{c}{Average} \\ 
& mIoU $\uparrow$ & mAcc $\uparrow$ & mIoU $\uparrow$ & mAcc $\uparrow$ & mIoU $\uparrow$ & mAcc $\uparrow$ & mIoU $\uparrow$ & mAcc $\uparrow$ & mIoU $\uparrow$ & mAcc $\uparrow$ & mIoU $\uparrow$ & mAcc $\uparrow$ & mIoU $\uparrow$ & mAcc $\uparrow$ \\  
\midrule[0.5pt]
LERF   \cite{LERF}   & 6.302  & 33.592 & 5.578  & 24.181 & 6.604  & 20.152 & 0.435  & 1.568  & 0.822  & 4.847  & 6.130  & 18.140 & 4.312 & 17.080 \\
3DOVS  \cite{3DOVS}  & 6.723  & 35.868 & 6.374  & 35.095 & 7.543  & 23.014 & 0.542  & 1.807  & 1.289  & 6.000  & 4.845  & 14.353 & 4.553 & 19.356 \\
Feature 3DGS \cite{feature3dgs}  & 6.737  & 36.329  & 8.502  & 38.607 & 10.803  & 42.202  & 6.844  & 25.242 & 9.190  & 34.651  & 15.428 & 52.440 & 9.584 & 38.245 \\
LangSplat  \cite{langsplat}      & 2.518  & 15.338  & 4.654 & 24.760 & 4.469  & 22.782  & 1.858 & 7.469  & 3.591  & 13.013  & 5.349 & 22.837 & 3.740 & 17.700 \\ 
\midrule[0.5pt]
Ours 
& \textbf{13.952} & \textbf{57.019} & \textbf{27.632} & \textbf{76.072} 
& \textbf{30.761} & \textbf{76.606} & \textbf{11.432} & \textbf{29.752} 
& \textbf{32.612} & \textbf{87.034} & \textbf{28.494} & \textbf{76.444} 
& \textbf{24.147} & \textbf{67.154} \\ 
\Xhline{3\arrayrulewidth}
\end{tabular}
\end{adjustbox}
\end{table*}

\begin{figure*}[!t]
\centering
\includegraphics[width=.95\linewidth]{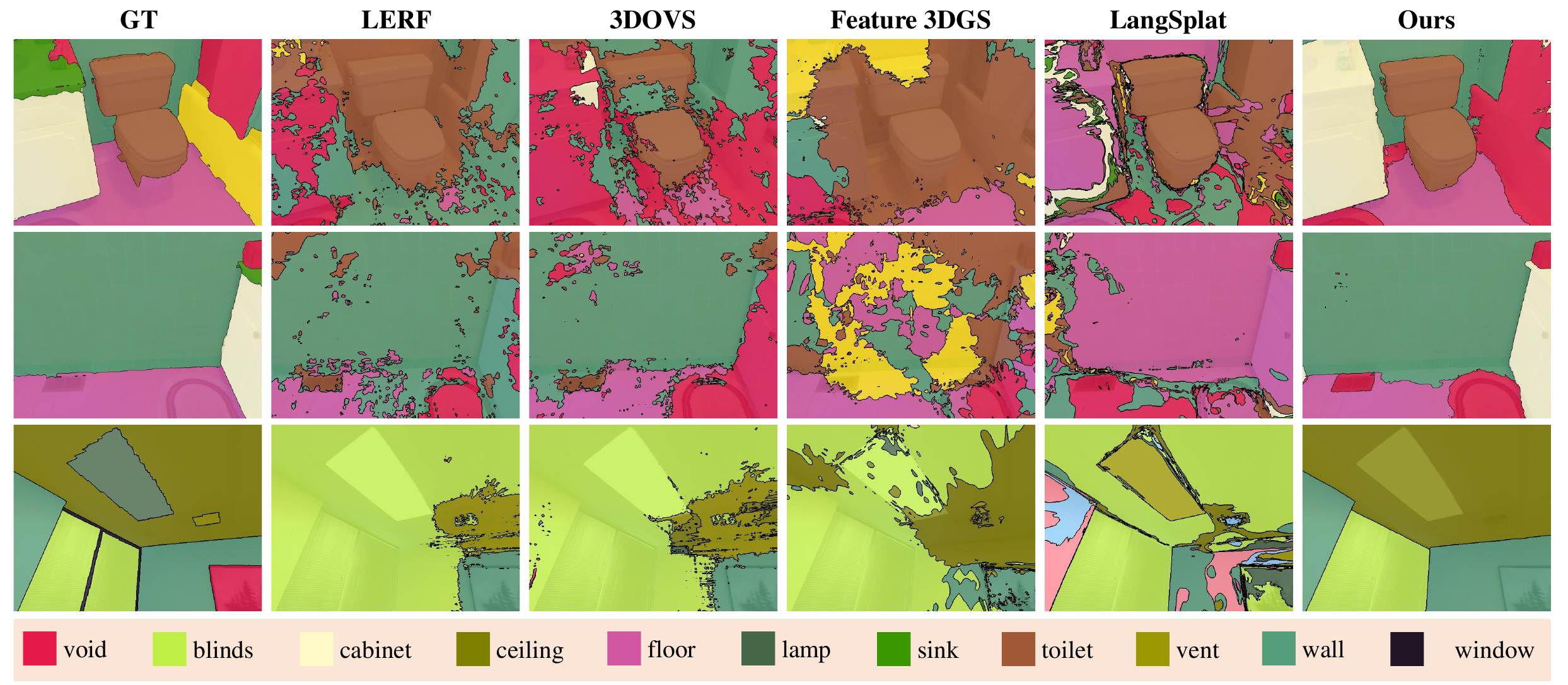}
\caption{
Visual segmentation results of novel view, under sparse-view training data conditions. Our approach obtains more robust and accurate segmentation performance. 
}
\label{fig:Vis_sparseview_seg}
\end{figure*}

\begin{table*}[t]
\renewcommand{\arraystretch}{1.2}
\caption{Performance comparison of segmentation results in novel views from the ScanNet dataset \cite{scannet}, utilizing multi-view training data and CLIP-LSeg \cite{LSeg} to optimize the Gaussian radiance field. 
Our proposed approach achieves superior results. 
} 
\label{table:scene_multieview_scannet_lseg}
\centering 
\begin{adjustbox} {width=.9\linewidth}
\begin{tabular}{ r | cc | cc | cc | cc | cc }
\Xhline{3\arrayrulewidth}
\multirow{2}{*}{Method} & \multicolumn{2}{c|}{Scene0004} & \multicolumn{2}{c|}{Scene0389} & \multicolumn{2}{c|}{Scene0494} & \multicolumn{2}{c|}{Scene0693} & \multicolumn{2}{c}{Average} \\ 
& mIoU $\uparrow$ & mAcc $\uparrow$ & mIoU $\uparrow$ & mAcc $\uparrow$ & mIoU $\uparrow$ & mAcc $\uparrow$ & mIoU $\uparrow$ & mAcc $\uparrow$ & mIoU $\uparrow$ & mAcc $\uparrow$  \\  
\midrule[0.5pt]
Feature 3DGS \cite{feature3dgs}  & 39.595 & 85.641 & 41.677 & 88.598 & 50.353 & 85.575 & 48.942 & 76.832 & 45.142 & 84.162 \\
Ours   
& \textbf{48.051} & \textbf{88.727} & \textbf{47.697} & \textbf{92.773} 
& \textbf{56.136} & \textbf{89.087} & \textbf{60.801} & \textbf{80.863} 
& \textbf{53.171} & \textbf{87.862} \\ 
\Xhline{3\arrayrulewidth}
\end{tabular}
\end{adjustbox}
\end{table*}

\begin{table*}[t]
\renewcommand{\arraystretch}{1.2}
\caption{Performance comparison of segmentation results in novel views from the Replica dataset \cite{replica}, utilizing multi-view training data and CLIP-LSeg \cite{LSeg} to optimize the Gaussian radiance field. 
Our proposed approach maintains superior results. 
} 
\label{table:scene_multieview_replica_lseg}
\centering 
\begin{adjustbox} {width=\linewidth}
\begin{tabular}{ r | cc | cc | cc | cc | cc | cc | cc }
\Xhline{3\arrayrulewidth}
\multirow{2}{*}{Method} & \multicolumn{2}{c|}{Room0} & \multicolumn{2}{c|}{Room1} & \multicolumn{2}{c|}{Room2} & \multicolumn{2}{c|}{Office0} & \multicolumn{2}{c|}{Office2} & \multicolumn{2}{c|}{Office4} & \multicolumn{2}{c}{Average} \\ 
& mIoU $\uparrow$ & mAcc $\uparrow$ & mIoU $\uparrow$ & mAcc $\uparrow$ & mIoU $\uparrow$ & mAcc $\uparrow$ & mIoU $\uparrow$ & mAcc $\uparrow$ & mIoU $\uparrow$ & mAcc $\uparrow$ & mIoU $\uparrow$ & mAcc $\uparrow$ & mIoU $\uparrow$ & mAcc $\uparrow$ \\  
\midrule[0.5pt]
Feature 3DGS \cite{feature3dgs}   & 25.169 & 71.148 & 25.297 & 72.115 & 29.746 & 80.812 & 24.226 & 65.128 & 34.516 & 88.565 & 33.743 & 81.298 & 28.783 & 76.514 \\ 
Ours 
& \textbf{26.043} & \textbf{73.994} & \textbf{27.366} & \textbf{74.608} 
& \textbf{30.314} & \textbf{82.814} & \textbf{35.263} & \textbf{78.388} 
& \textbf{39.315} & \textbf{90.007} & \textbf{41.097} & \textbf{82.374}  
& \textbf{33.233} & \textbf{80.364} \\ 
\Xhline{3\arrayrulewidth}
\end{tabular}
\end{adjustbox}
\end{table*}

\begin{figure*}[!t]
\centering
\includegraphics[width=\linewidth]{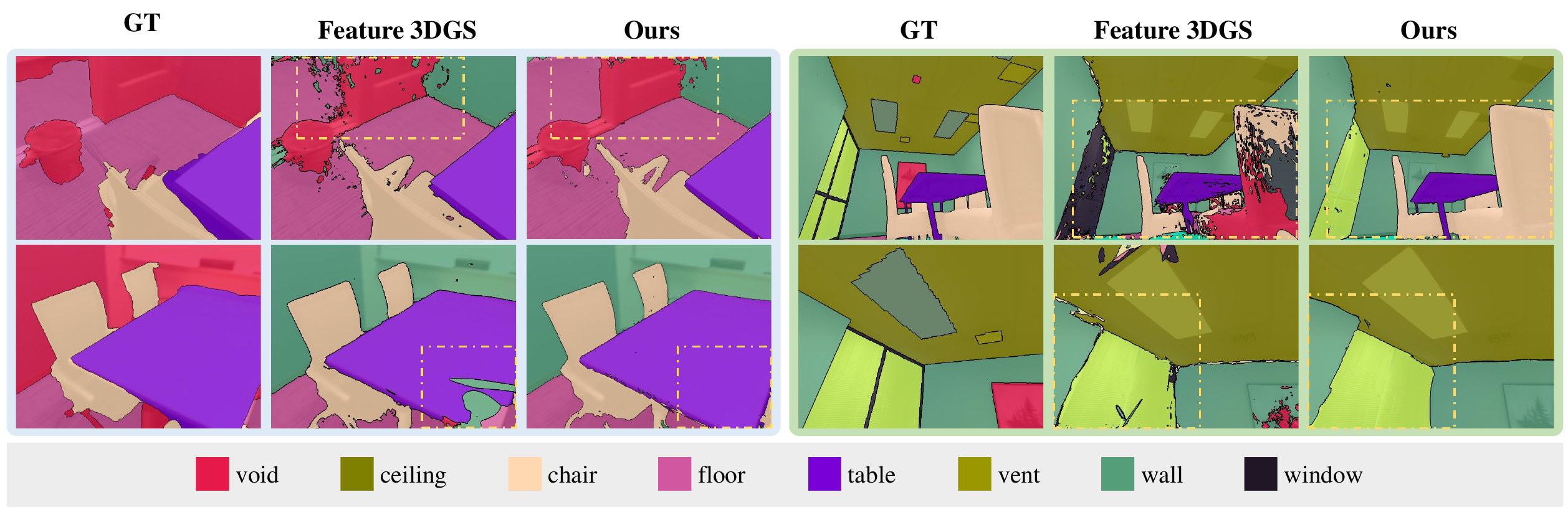}
\caption{
Visual segmentation results of novel views using CLIP-LSeg \cite{LSeg} for semantic Gaussian attribute optimization. 
Our approach demonstrates more accurate and coherent segmentation results. 
}
\label{fig:Vis_lseg_seg}
\end{figure*}

\subsubsection{\textbf{Results on the Sparse-view Benchmark Datasets}} 
We introduce a sparse-view evaluation benchmark, a setup that trains the model solely using sparse training views, to evaluate the robustness of methods. 

We report the sparse-view quantitative comparisons in Table \ref{table:scene_sparseview_scannet} and \ref{table:scene_sparseview_replica}. 
Our approach, despite being trained using sparse views, consistently achieves superior performance in segmentation precision. 
Concretely, our method surpasses the second-best method by 23.66\% and 14.56\% in mIoU on ScanNet and Replica, respectively. 
Existing methods fail to produce accurate semantic results, because they suffer from insufficient view-consistent constraints, making them sensitive to sparse inputs. 
Moreover, LangSplat's approach of fixing the position, scaling, and rotation of 3D Gaussians during semantic parameter optimization inhibits flexible adjustment of semantic representations, further impeding accuracy.

Fig. \ref{fig:Vis_sparseview_seg} presents visual comparisons of various methods under sparse and under-constrained conditions.  
Specifically, the 3DOVS method tends to produce ambiguous visual results, and Feature 3DGS exhibits coarse semantic outputs due to the lack of sufficient semantic-consistent constraints under the sparse input conditions. 
Moreover, LangSplat encounters significant difficulties in sparse-input scenarios. 
This limitation arises because, during semantic parameter optimization, LangSplat inherits the inferior Gaussian representations (including position, scaling, and rotation) obtained from sparse-input scene reconstruction using vanilla 3DGS, and subsequently optimizes only semantic parameters. Consequently, these optimization constraints lead to suboptimal semantic understanding performance.  
In contrast, our method shows a more complete structure across various scenes. 
This can be attributed to our 3D coherent regularization strategy, which effectively integrates semantic information from adjacent views and enforces 3D consistency constraints, ensuring robustness even with sparse inputs.

\begin{table*}[t]
\renewcommand{\arraystretch}{1.2}
\caption{Quantitative comparison on reconstruction results of novel views in Replica and ScanNet datasets.
Our proposed approach maintains reconstruction quality on par with the 3D Gaussian Splatting (3DGS) method \cite{3dgs}. 
} 
\label{table:SOTA_reconstruction}
\centering 
\begin{adjustbox} {width=\linewidth}
\begin{tabular}{ c | ccc | ccc | ccc | ccc }
\Xhline{3\arrayrulewidth}
\multirow{3}{*}{Method} & \multicolumn{6}{c|}{\textit{Multi-view Training Data}} & \multicolumn{6}{c}{\textit{Sparse-view Training Data}} \\ 
~ & \multicolumn{3}{c}{Replica \cite{replica}} & \multicolumn{3}{c|}{ScanNet \cite{scannet}} & \multicolumn{3}{c}{Replica \cite{replica}} & \multicolumn{3}{c}{ScanNet \cite{scannet}}\\ 
& PSNR $\uparrow$ & SSIM $\uparrow$ & \multicolumn{1}{c}{LPIPS $\downarrow$ } 
& PSNR $\uparrow$ & SSIM $\uparrow$ & LPIPS $\downarrow$ 
& PSNR $\uparrow$ & SSIM $\uparrow$ & \multicolumn{1}{c}{LPIPS $\downarrow$ } 
& PSNR $\uparrow$ & SSIM $\uparrow$ & LPIPS $\downarrow$ \\ 
\hline
LERF   \cite{LERF}    & 31.034 & 0.904 & 0.103  & 24.921  & 0.778  & 0.392   & 17.509 & 0.697 & 0.484 & 20.819 & 0.719 & 0.429  \\ 
3DOVS  \cite{3DOVS}   & 31.373 & 0.908 & 0.091  & 24.915  & 0.780  & 0.389   & 17.923 & 0.708 & 0.477 & 21.414 & 0.723 & 0.422  \\ 
\hline
3DGS   \cite{3dgs}    & 35.477 & 0.955 & 0.090  & 28.265  & 0.857  & 0.258   & 26.333 & 0.880 & 0.196  & 22.713 & 0.752 & 0.354  \\ 
Feature 3DGS  \cite{feature3dgs}   & 35.439 & 0.955 & 0.090  & 28.453  & 0.863  & 0.250   & 26.313 & 0.881 & 0.193  & 22.224 & 0.741 & 0.355  \\ 
LangSplat  \cite{langsplat}        & 35.490 & 0.955 & 0.090  & 28.441  & 0.860  & 0.257   & 26.247 & 0.881 & 0.195  & 22.125 & 0.751 & 0.354  \\  
\hline
Ours    & 35.519 & 0.955 & 0.090 & 29.021 & 0.866 & 0.250  & 26.662 & 0.885 & 0.191 & 22.760 & 0.759  & 0.350  \\ 
\Xhline{3\arrayrulewidth}
\end{tabular}
\end{adjustbox}
\end{table*}

\begin{figure*}[!t]
\centering
\includegraphics[width=.99\linewidth]{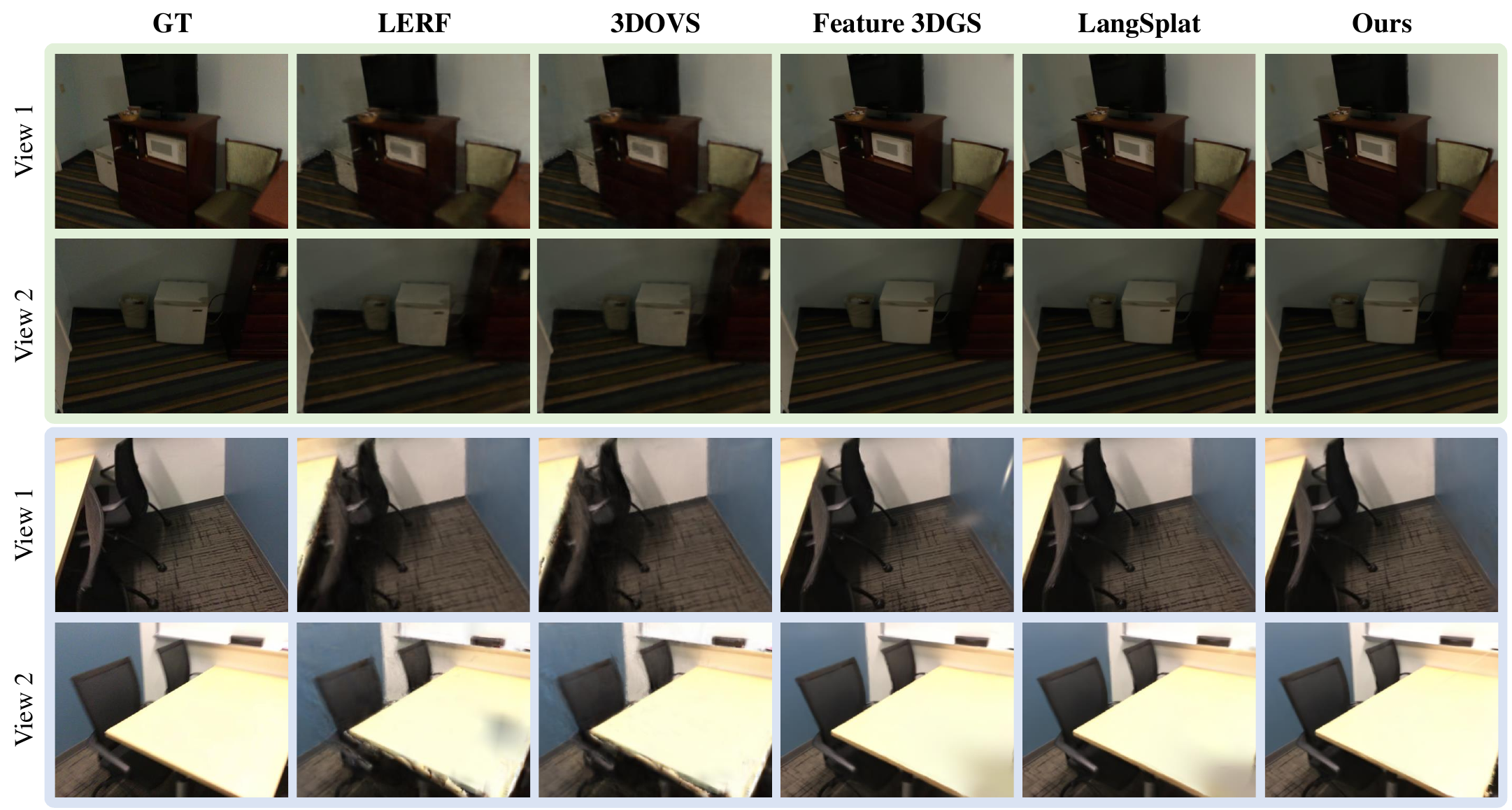}
\caption{
Visual reconstruction results of novel view, under multi-view training data conditions. 
Our method achieves photo-realistic rendering quality comparable to the Gaussian-based method, outperforming NeRF-based approaches. 
}
\label{fig:Vis_multiview_rgb}
\end{figure*}

\subsubsection{\textbf{Results using CLIP-LSeg Model}} 
We further present experimental results using various vision-language foundation models, such as CLIP-LSeg \cite{LSeg}, which was previously used in by Feature 3DGS \cite{feature3dgs} for optimizing Gaussian semantic attributes. 
The quantitative results presented in Table \ref{table:scene_multieview_scannet_lseg} and \ref{table:scene_multieview_replica_lseg}, indicate that our approach consistently achieves superior segmentation results across diverse scenes. 
Additionally, the visual comparison results shown in Fig. \ref{fig:Vis_lseg_seg}, demonstrate that Feature 3DGS easily produces ambiguous semantics due to a lack of view-consistent constraints, whereas our approach delivers more accurate and coherent segmentation results. 
These results underscore the effectiveness and generalizability of our method under different CLIP models.

\subsubsection{\textbf{Scene Reconstruction Results}} 
We also report the reconstruction quality to demonstrate that our method not only delivers accurate segmentation results, but also achieves high-quality scene reconstruction. 

\begin{figure*}[!t]
\centering
\includegraphics[width=.97\linewidth]{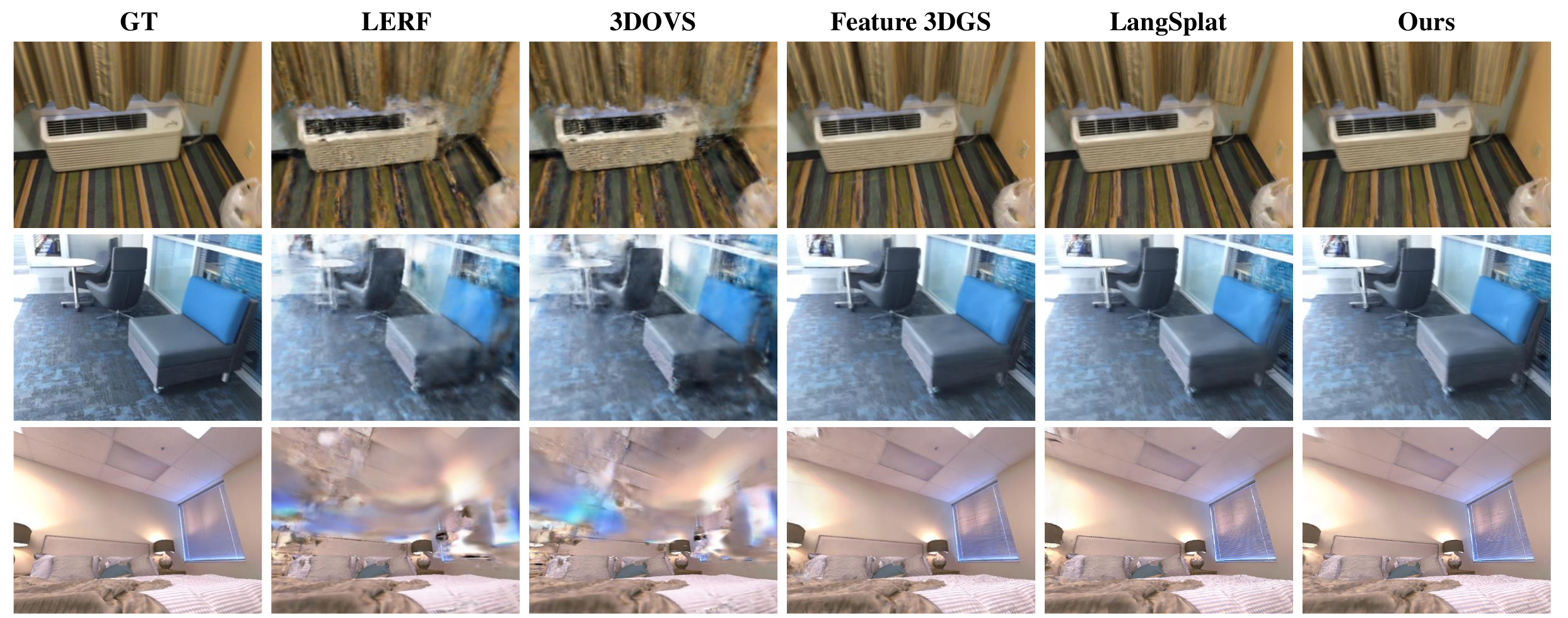}
\caption{
Visual reconstruction results of novel view, under sparse-view training data conditions. Our approach shows robust reconstruction quality across various scenes. 
}
\label{fig:Vis_sparseview_rgb}
\end{figure*}

As shown in Table \ref{table:SOTA_reconstruction}, our method achieves reconstruction quality on par with other 3DGS-based methods and suppresses NeRF-based approaches under both multi-view and sparse-view training conditions
Furthermore, we provide qualitative results in Fig. \ref{fig:Vis_multiview_rgb} and Fig. \ref{fig:Vis_sparseview_rgb}. We can observe that our approach consistently renders photo-realistic details across diverse synthetic and real-world scenes. 
In summary, these results demonstrate the effectiveness of our method in simultaneously reconstructing and semantically understanding 3D scenes.

\subsubsection{\textbf{Efficiency Comparison}} \label{sec_efficiency}
We report the training time and inference results for efficiency comparison, as summarized in Table \ref{table:efficiency_comparsion}. All evaluations are conducted using a single NVIDIA A100 GPU. 
The evaluation highlights the superior efficiency of our approach, which outperforms competing methods in both training and inference speed.
Specifically, NeRF-based methods suffer from slow speeds due to the computational demands of volume rendering.  
Although Feature 3DGS adopts a fast splatting technique, it faces an efficiency bottleneck caused by embedding high-dimensional semantic features into 3D Gaussians, which particularly hampers training efficiency. 
Similarly, LangSplat's efficiency is limited by its time-consuming autoencoder pre-training and post-processing upsampling processes.   
In contrast, our approach achieves superior efficiency by modeling compact semantic Gaussian representations through our SAC strategy, enabling a faster rendering speed (\textgreater100 FPS).

\begin{table*}[t]
\renewcommand{\arraystretch}{1.1}
\caption{Comparison with current methods on training time and inference speed. More details refer to Sec. \ref{sec_efficiency}. 
} 
\label{table:efficiency_comparsion}
\centering 
\large
\begin{adjustbox} {width=.82\linewidth}
\begin{tabular}{ r | p{3.4cm}<{\centering} p{3.4cm}<{\centering} | p{3.4cm}<{\centering} p{3.4cm}<{\centering} p{3.4cm}<{\centering} }
\Xhline{3\arrayrulewidth}
\multirow{2}{*}{Method} & \multicolumn{2}{c|}{NeRF-based} & \multicolumn{3}{c}{3DGS-based} \\ 
~ & LERF \cite{LERF} & 3DOVS \cite{3DOVS} & Feature 3DGS \cite{feature3dgs} & LangSplat \cite{langsplat} & Ours  \\  
\midrule[0.5pt]
Training Time $\downarrow$  & $\sim$ 2h 10 mins  & $\sim$ 2h & $\sim$ 14h 50 mins & $\sim$ 1h 24mins & \textbf{$\sim$ 20mins} \\ 
Inference FPS $\uparrow$    & 0.2  & 0.3 & 2.5 & 45 & \textbf{190} \\ 
\Xhline{3\arrayrulewidth}
\end{tabular}
\end{adjustbox}
\end{table*}

\begin{table*}[!t]
\renewcommand{\arraystretch}{1.1}
\caption{
Ablation studies for our approach. 
SAC: Semantic Attribute Compactness. 3DCR: 3D Coherent Regularization. PDR: Progressive Densification Regulation. 
"w/o coherent" denotes excluding semantic information integration from adjacent views. 
Settings (b1)-(b2) assess the impact of different feature dimensions of the semantic parameter, while (c1)-(c4) evaluate the effect of introducing 3DCR at various stages of training.
The final configuration is denoted by \textbf{Bold}.  
}
\label{table_Ablation_all}
\centering
\normalsize
\setlength{\tabcolsep}{0.9mm}{
\begin{tabular}{ p{1.2cm}<{\centering} | c | p{1.0cm}<{\centering} | p{1.4cm}<{\centering} p{1.4cm}<{\centering} | p{1.4cm}<{\centering} p{1.4cm}<{\centering} }
\toprule[1pt] 
\multirow{2}{*}{Index} & \multirow{2}{*}{Setting} & \multirow{2}{*}{FPS}  
& \multicolumn{2}{c|}{Room0} & \multicolumn{2}{c}{Scene0494} \\
&  &  &  mIoU $\uparrow$ & mAcc $\uparrow$  & mIoU $\uparrow$ & mAcc $\uparrow$ \\ 
\midrule[0.8pt]
(a)   & Baseline        &  2.5  & 8.794   & 39.916  & 25.789  & 52.006 \\ 
(b)   & (a) + SAC       &  150  & 19.920  & 63.125  & 36.480  & 73.529 \\
(c)   & (b) + 3DCR ($\mathcal{L}_{consistency}^{2D}$) &  150  & 25.564  & 71.918  & 54.023  & 85.636 \\ 

(d)   & (b) + 3DCR ($\mathcal{L}_{consistency}^{3D}$)   & 150  &  23.986 & 68.810  & 53.750 & 83.900 \\ 

(e)   & (b) + 3DCR ($\mathcal{L}_{consistency}^{2D}$, $\mathcal{L}_{consistency}^{3D}$) &  150  & 26.852  & 73.542  & 58.293  & 87.366 \\ 
(f)   & (e) + PDR (Ours) &  190  & \textbf{27.249} & \textbf{74.129}  & \textbf{58.543}  & \textbf{88.224} \\ 

\midrule[0.5pt] 
(g)   & Ours w/o coherent   & 190  &  25.445 & 71.564 & 56.633 & 86.863  \\ 
(h)   & Ours w 3DCR (0.5 * $\mathcal{L}_{consistency}^{3D}$)   & {190}  & {27.055}  & {74.044}  & {58.483}  & {88.037}  \\  
{(i)}   & {Ours w 3DCR (1.5 * $\mathcal{L}_{consistency}^{3D}$)}   & {190}  & {27.739} 	& {74.166}  & {58.152}  & {88.005}  \\ 

\midrule[0.5pt]
(b1)  & Ours ($d$=3)    & 190  & \textbf{27.249} & \textbf{74.129} & \textbf{58.543}  & \textbf{88.224} \\ 
(b2)  & Ours ($d$=8)    & 175  & 27.888  & 72.995  & 58.729 & 88.174  \\ 
\midrule[0.5pt]
(c1)  & Ours ($\mathcal{T}$=10k) &  190  & 24.794  & 71.036  & 57.810 & 87.467 \\ 
(c2)  & Ours ($\mathcal{T}$=15k) &  190  & \textbf{27.249} & \textbf{74.129}  & \textbf{58.543}  & \textbf{88.224} \\ 
(c3)  & Ours ($\mathcal{T}$=20k) &  190  & 26.034  & 73.068  & 57.585 & 86.801 \\ 
(c4)  & Ours ($\mathcal{T}$=25k) &  190  & 24.309  & 70.265  & 56.561 & 86.007 \\ 
\bottomrule[1pt]
\end{tabular}}
\end{table*}

\begin{figure*}[!t]
\centering
\includegraphics[width=.8\linewidth]{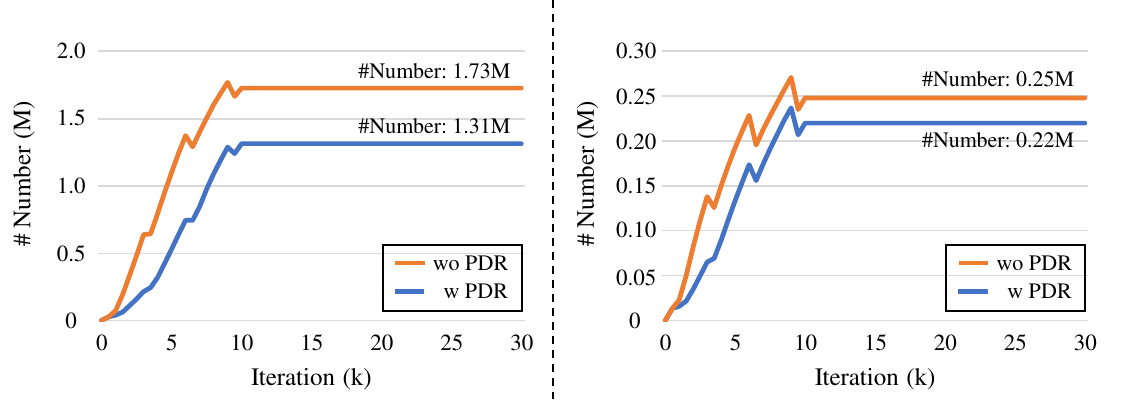}
\caption{Visualization of the varying count of Gaussians in the Room0 scene (left) and Scene0494 scene (right) during training. 
The vertical axis indicates the number of Gaussians ("\# Number"), while the horizontal axis represents the training iteration ("\# Iteration"). 
These results show that PDR effectively regulates the number of Gaussian, improving rendering efficiency. 
}
\label{fig:Vis_PDR}
\end{figure*}

\begin{figure*}[!t]
\centering
\includegraphics[width=.75\linewidth]{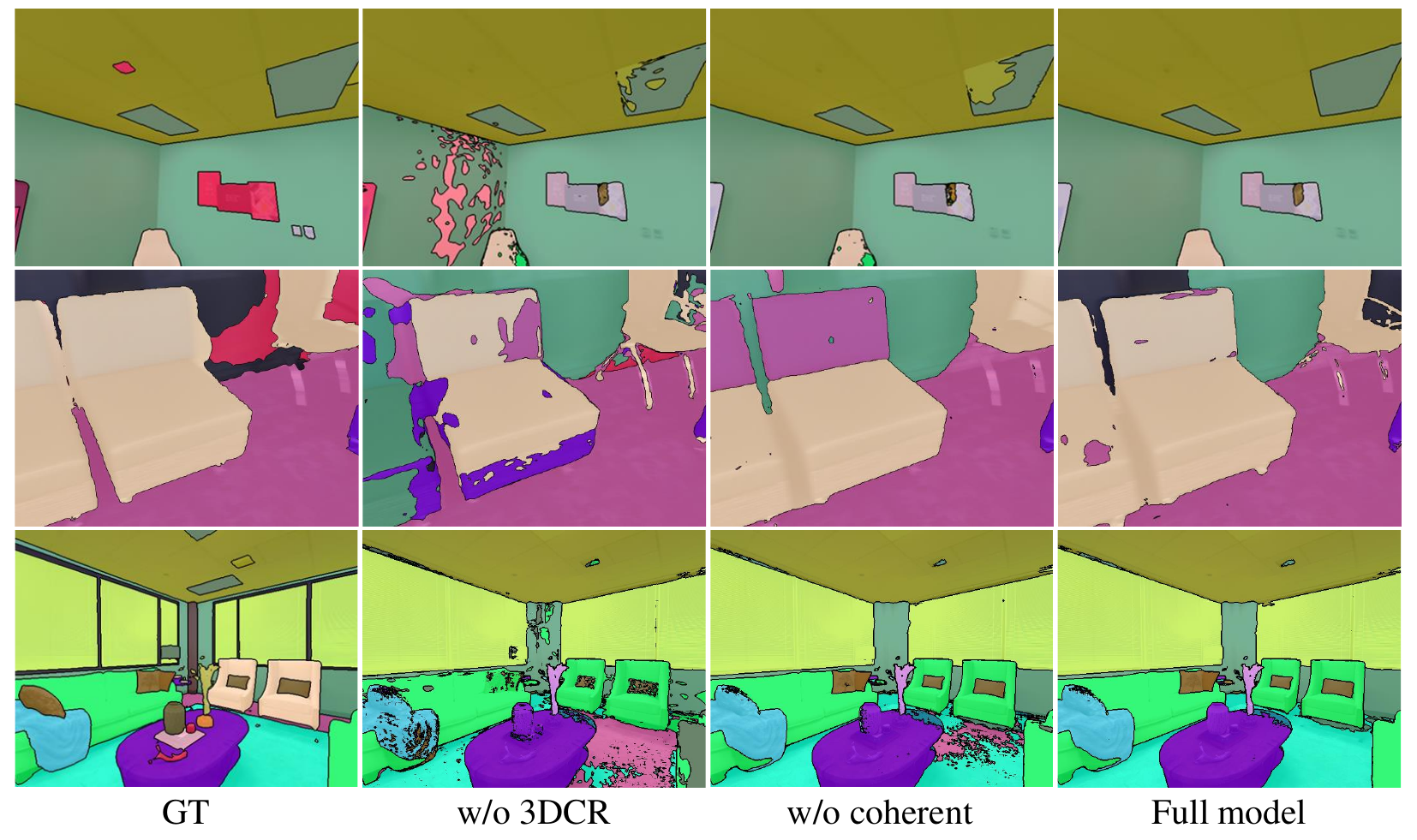}
\caption{
Visual comparison of ablation experiments. 
As illustrated, incorporating the 3D coherent regularization (2$^{nd}$ column to 4$^{th}$ column) effectively mitigates segmentation ambiguities and improves coherence. 
Besides, the integration of semantic information from adjacent views for regularization (3$^{rd}$ column to 4$^{th}$ column) enhances object completeness, leading to more precise segmentation results.   
}
\label{fig:vis_ablation}
\end{figure*}

\subsection{Ablation Studies} \label{section:ablation_study}
In this section, we present ablation experiments to evaluate the effectiveness of each component in our approach.

\textbf{Effectiveness of proposed components.}
We adopt Feature 3DGS \cite{feature3dgs} as the experimental baseline. 
As shown in Table \ref{table_Ablation_all}, we gradually employ different components, including Semantic Attribute Compactness (SAC), 3D Coherent Regularization (3DCR), and Progressive Densification Regulation (PDR), to evaluate their effectiveness. 


\begin{itemize}
    \item Compared to baseline (a), adding SAC significantly improves inference efficiency. This enhancement is due to SAC's capacity to embed compact semantic information into 3D Gaussians. Moreover, SAC's region-level unified processing refines object boundaries, leading to more precise segmentation results.

    \item 
    As exhibited in (c) and (d), further applying 3DCR achieves notable performance improvements without adding extra computational overhead during inference. 
    Moreover, combining both 2D and 3D constraints, as demonstrated in (e), leads to better performance, highlighting their complementary nature.
    These results validate that 3DCR introduces essential semantic-consistent constraints, effectively mitigating ambiguous and coarse semantics and improving the overall coherence of the segmentation results. 
    We further analyze the impact of different regularization weights on the 3DCR loss. As shown in (h) and (i), both weaker and stronger regularization weights maintain robust performance, demonstrating the stability and effectiveness of our 3DCR mechanism across a range of settings.

    \item As shown in (e), incorporating PDR further boosts computational efficiency and delivers a modest accuracy improvement. 
    This improvement stems from PDR's ability to regulate the quantity of Gaussian primitives, which not only accelerates rasterization but also facilitates more effective semantic Gaussian radiance field construction. 
    Moreover, Fig. \ref{fig:Vis_PDR} also highlights the effectiveness of PDR in regulating Gaussian quantities. 
\end{itemize}

Additionally, we present visual ablation results in Fig. \ref{fig:vis_ablation}. 
It can be seen that integrating the 3D coherent regularization significantly reduces semantic ambiguity and enhances consistency compared to removing it ("w/o 3DCR"). 
Furthermore, incorporating semantic information from adjacent views improves scene coherence compared to omitting it ("w/o coherent"). 
These results show the effectiveness of our proposed strategies in achieving precise and coherent 3D semantic understanding.

\begin{figure*}[!t]
\centering
\includegraphics[width=.78\linewidth]{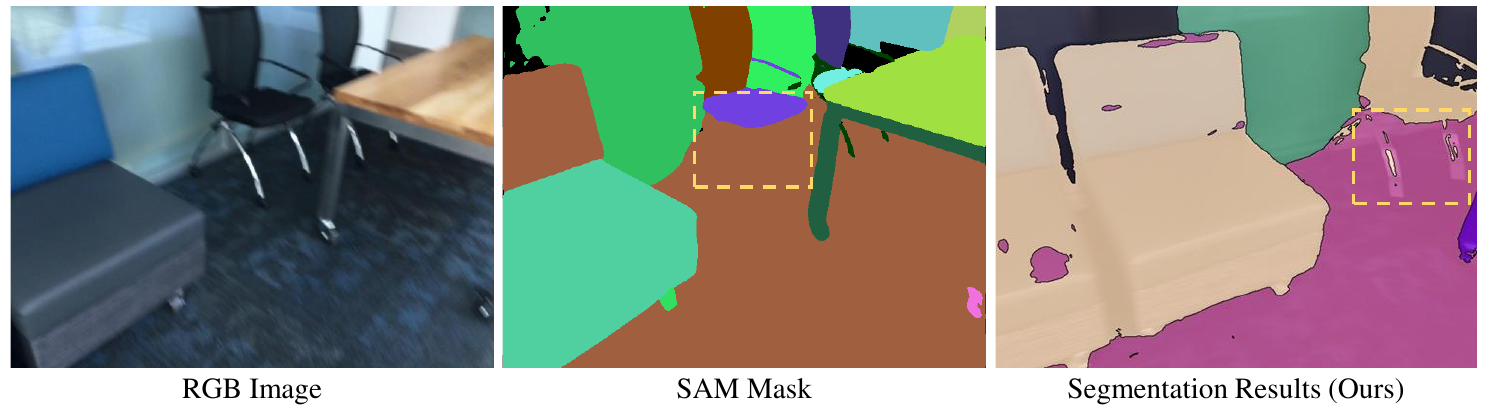}
\caption{
{
Illustration of a SAM failure case caused by motion blur. The chair leg (yellow dashed box) is incorrectly merged with the floor region in the generated mask, leading to imprecise segmentation. 
}
}
\label{fig:Vis_analysis_sam}
\end{figure*}

\begin{figure*}[!t]
\centering
\includegraphics[width=.95\linewidth]{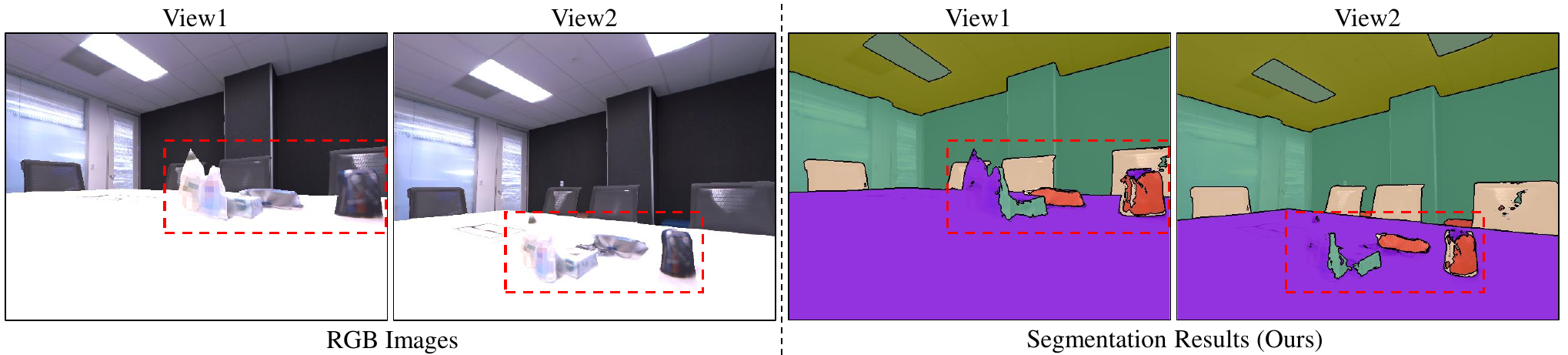}
\caption{
{
A failure case in a cluttered, low-contrast region (red dashed box), where weak CLIP features result in inaccurate semantic predictions. 
}
}
\label{fig:Vis_analysis_clip}
\end{figure*}

\textbf{Effectiveness of coherent in 3DCR.} 
As shown in (f) of Table \ref{table_Ablation_all}, "w/o coherent" denotes the absence of semantics from adjacent views for regularization. Instead, the semantics are only regularized using the current view's masks from SAM. The regularized semantics are then applied as ${Z}$ in Equation \eqref{eq_2d_consistency}. 
The decrease in performance verifies that incorporating semantics from adjacent views improves semantic consistency.

\textbf{Analysis of $d$ in SAC.}
We analyze the impact of different $d$ in SAC, illustrated in (b1) - (b2).
As $d$ increases, the inference speed declines, coupled with a marginal improvement in mIoU. Therefore, we opt for $d=3$ as it delivers efficient yet competitive results.

\textbf{Analysis of $\mathcal{T}$ in 3DCR.}
We examine the influence of different settings of $\mathcal{T}$, as depicted in (c1) - (c4). Specifically, $\mathcal{T}=10k$ denotes the application of our 3D coherent regularization strategy after 10,000 training iterations. 
When $\mathcal{T}$ is set to 15k, the model achieves overall superior results, making it the chosen final configuration.

\subsection{Discussion}
\textbf{Real-World Applications Potential.}
The proposed CLIP-GS, with its efficient, view-consistent, language-driven semantic understanding capability, holds significant promise for various real-world applications. 
In robotics, precise 3D scene understanding and text-driven object segmentation can improve navigation, manipulation, and grasp planning while eliminating the need for costly manual annotations. 
In augmented and virtual reality, CLIP-GS enables scene understanding and natural language querying, supporting immersive interaction and context-aware content placement. 
In summary, CLIP-GS paves the way for more intelligent and responsive 3D applications in robotics and AR/VR environments. 

\noindent
\textbf{Limitations.}   
While our approach achieves precise and efficient 3D semantic understanding, we acknowledge potential limitations caused by the performance of 2D foundation models, such as SAM and CLIP. 
On the one hand, the quality of object-level region masks generated by SAM may occasionally be imperfect in some complex indoor environments, affecting the results of our method. 
For example, as shown in Fig. \ref{fig:Vis_analysis_sam}, motion blur in the input image can cause a chair’s leg to be incorrectly merged with the floor, resulting in inaccurate mask generation. Such imperfect masks may affect the semantic learning of 3D Gaussians and lead to incomplete segmentation in novel views. Future work may explore enhanced region segmentation strategies or adaptive mask refinement techniques to mitigate this issue.
On the other hand, as illustrated in Fig. \ref{fig:Vis_analysis_clip}, our method encounters challenges in cluttered or low-contrast regions, where accurate object segmentation becomes difficult. This limitation arises because the semantic supervision relies on CLIP features, which may be weaker in such cases. Future work will integrate more robust vision-language foundation models to enhance robustness in challenging scenarios.

\section{Conclusion}
\label{sec:conclusion}
In this work, we present CLIP-GS, a novel approach utilizing CLIP-informed 3D Gaussian Splatting for achieving real-time, view-consistent 3D semantic understanding in indoor Scenes. 
CLIP-GS incorporates Semantic Attribute Compactness (SAC) to embed compact semantic information into 3D Gaussians, allowing for highly efficient rendering. 
Additionally, the proposed 3D Coherent Regularization (3DCR) enhances semantic-consistent constraints across multiple viewpoints, facilitating more coherent 3D segmentation. 
Experimental results demonstrate that our approach significantly outperforms SOTA methods on synthetic and real-world indoor scenes. Furthermore, our method maintains superior performance even under sparse input conditions, substantiating its robustness.

%% file: main.bbl
\begin{thebibliography}{58}
\providecommand{\natexlab}[1]{#1}
\providecommand{\url}[1]{\texttt{#1}}
\expandafter\ifx\csname urlstyle\endcsname\relax
  \providecommand{\doi}[1]{doi: #1}\else
  \providecommand{\doi}{doi: \begingroup \urlstyle{rm}\Url}\fi

\bibitem[Bao et~al.(2025)Bao, Ding, Huo, Liu, Li, Li, Gao, and Luo]{bao20253d}
Yanqi Bao, Tianyu Ding, Jing Huo, Yaoli Liu, Yuxin Li, Wenbin Li, Yang Gao, and Jiebo Luo.
\newblock 3d gaussian splatting: Survey, technologies, challenges, and opportunities.
\newblock \emph{IEEE Transactions on Circuits and Systems for Video Technology}, 2025.

\bibitem[Barron et~al.(2023)Barron, Mildenhall, Verbin, Srinivasan, and Hedman]{zipnerf}
Jonathan~T. Barron, Ben Mildenhall, Dor Verbin, Pratul~P. Srinivasan, and Peter Hedman.
\newblock Zip-nerf: Anti-aliased grid-based neural radiance fields.
\newblock In \emph{Proceedings of the IEEE/CVF International Conference on Computer Vision}, pages 19697--19705, 2023.

\bibitem[Chen et~al.(2022)Chen, Xu, Geiger, Yu, and Su]{tensorf}
Anpei Chen, Zexiang Xu, Andreas Geiger, Jingyi Yu, and Hao Su.
\newblock Tensorf: Tensorial radiance fields.
\newblock In \emph{Proceedings of the European Conference on Computer Vision}, pages 333--350, 2022.

\bibitem[Cheng et~al.(2023)Cheng, Oh, Price, Schwing, and Lee]{deva}
Ho~Kei Cheng, Seoung~Wug Oh, Brian Price, Alexander Schwing, and Joon-Young Lee.
\newblock Tracking anything with decoupled video segmentation.
\newblock In \emph{Proceedings of the IEEE/CVF International Conference on Computer Vision}, pages 1316--1326, 2023.

\bibitem[Cherti et~al.(2023)Cherti, Beaumont, Wightman, Wortsman, Ilharco, Gordon, Schuhmann, Schmidt, and Jitsev]{OpenCLIP}
Mehdi Cherti, Romain Beaumont, Ross Wightman, Mitchell Wortsman, Gabriel Ilharco, Cade Gordon, Christoph Schuhmann, Ludwig Schmidt, and Jenia Jitsev.
\newblock Reproducible scaling laws for contrastive language-image learning.
\newblock In \emph{Proceedings of the IEEE/CVF Conference on Computer Vision and Pattern Recognition}, pages 2818--2829, 2023.

\bibitem[Choi et~al.(2024)Choi, Song, Kim, Kim, and Do]{clickgaussian}
Seokhun Choi, Hyeonseop Song, Jaechul Kim, Taehyeong Kim, and Hoseok Do.
\newblock Click-gaussian: Interactive segmentation to any 3d gaussians.
\newblock In \emph{Proceedings of the European Conference on Computer Vision}, pages 289--305. Springer, 2024.

\bibitem[Chu and Harada(2024)]{chu2024generalizable}
Xuangeng Chu and Tatsuya Harada.
\newblock Generalizable and animatable gaussian head avatar.
\newblock In \emph{Proceedings of the Advances in Neural Information Processing Systems}, pages 57642--57670, 2024.

\bibitem[Dai et~al.(2017{\natexlab{a}})Dai, Chang, Savva, Halber, Funkhouser, and Nie{\ss}ner]{scannet}
Angela Dai, Angel~X Chang, Manolis Savva, Maciej Halber, Thomas Funkhouser, and Matthias Nie{\ss}ner.
\newblock Scannet: Richly-annotated 3d reconstructions of indoor scenes.
\newblock In \emph{Proceedings of the IEEE conference on computer vision and pattern recognition}, pages 5828--5839, 2017{\natexlab{a}}.

\bibitem[Dai et~al.(2017{\natexlab{b}})Dai, Nie{\ss}ner, Zollh{\"o}fer, Izadi, and Theobalt]{Bundlefusion}
Angela Dai, Matthias Nie{\ss}ner, Michael Zollh{\"o}fer, Shahram Izadi, and Christian Theobalt.
\newblock Bundlefusion: Real-time globally consistent 3d reconstruction using on-the-fly surface reintegration.
\newblock \emph{ACM Transactions on Graphics}, 36\penalty0 (4):\penalty0 1, 2017{\natexlab{b}}.

\bibitem[Fei et~al.(2024)Fei, Xu, Zhang, Zhou, Yang, and He]{fei20243d}
Ben Fei, Jingyi Xu, Rui Zhang, Qingyuan Zhou, Weidong Yang, and Ying He.
\newblock 3d gaussian splatting as new era: A survey.
\newblock \emph{IEEE Transactions on Visualization and Computer Graphics}, 2024.

\bibitem[Fischer et~al.(2024)Fischer, Kulhanek, Bul{\`o}, Porzi, Pollefeys, and Kontschieder]{fischer2024dynamic}
Tobias Fischer, Jonas Kulhanek, Samuel~Rota Bul{\`o}, Lorenzo Porzi, Marc Pollefeys, and Peter Kontschieder.
\newblock Dynamic 3d gaussian fields for urban areas.
\newblock In \emph{Proceedings of the Advances in Neural Information Processing Systems}, pages 80466--80494, 2024.

\bibitem[Gao et~al.(2022)Gao, Gao, He, Lu, Xu, and Li]{nerfreview}
Kyle Gao, Yina Gao, Hongjie He, Dening Lu, Linlin Xu, and Jonathan Li.
\newblock Nerf: Neural radiance field in 3d vision, a comprehensive review.
\newblock \emph{arXiv preprint arXiv:2210.00379}, 2022.

\bibitem[Guo et~al.(2024)Guo, Ma, Fan, Liu, and Li]{guo2024semantic}
Jun Guo, Xiaojian Ma, Yue Fan, Huaping Liu, and Qing Li.
\newblock Semantic gaussians: Open-vocabulary scene understanding with 3d gaussian splatting.
\newblock \emph{arXiv preprint arXiv:2403.15624}, 2024.

\bibitem[Hu et~al.(2024)Hu, Wang, Fan, Fan, Peng, Lei, Li, and Zhang]{saga}
Xu Hu, Yuxi Wang, Lue Fan, Junsong Fan, Junran Peng, Zhen Lei, Qing Li, and Zhaoxiang Zhang.
\newblock Semantic anything in 3d gaussians.
\newblock \emph{arXiv preprint arXiv:2401.17857}, 2024.

\bibitem[Huang et~al.(2023)Huang, Mees, Zeng, and Burgard]{robotnavigation}
Chenguang Huang, Oier Mees, Andy Zeng, and Wolfram Burgard.
\newblock Visual language maps for robot navigation.
\newblock In \emph{Proceedings of the IEEE International Conference on Robotics and Automation}, pages 10608--10615, 2023.

\bibitem[In~Lee et~al.(2025)In~Lee, Park, Seo, Park, Park, Dam~Baek, Sangheon, Kim, et~al.]{in2025editsplat}
Dong In~Lee, Hyeongcheol Park, Jiyoung Seo, Eunbyung Park, Hyunje Park, Ha Dam~Baek, Shin Sangheon, Sangpil Kim, et~al.
\newblock Editsplat: Multi-view fusion and attention-guided optimization for view-consistent 3d scene editing with 3d gaussian splatting.
\newblock In \emph{Proceedings of the IEEE/CVF Conference on Computer Vision and Pattern Recognition}, 2025.

\bibitem[Jiao et~al.(2024)Jiao, Dong, Yin, Jie, Qian, Zhao, Shi, and Wei]{jiao2024clip}
Siyu Jiao, Haoye Dong, Yuyang Yin, Zequn Jie, Yinlong Qian, Yao Zhao, Humphrey Shi, and Yunchao Wei.
\newblock Clip-gs: Unifying vision-language representation with 3d gaussian splatting.
\newblock \emph{arXiv preprint arXiv:2412.19142}, 2024.

\bibitem[Kerbl et~al.(2023)Kerbl, Kopanas, Leimk{\"u}hler, and Drettakis]{3dgs}
Bernhard Kerbl, Georgios Kopanas, Thomas Leimk{\"u}hler, and George Drettakis.
\newblock 3d gaussian splatting for real-time radiance field rendering.
\newblock \emph{ACM Transactions on Graphics}, 42\penalty0 (4):\penalty0 1--14, 2023.

\bibitem[Kerr et~al.(2023)Kerr, Kim, Goldberg, Kanazawa, and Tancik]{LERF}
Justin Kerr, Chung~Min Kim, Ken Goldberg, Angjoo Kanazawa, and Matthew Tancik.
\newblock Lerf: Language embedded radiance fields.
\newblock In \emph{Proceedings of the IEEE/CVF International Conference on Computer Vision}, pages 19729--19739, 2023.

\bibitem[Kirillov et~al.(2023)Kirillov, Mintun, Ravi, Mao, Rolland, Gustafson, Xiao, Whitehead, Berg, Lo, Dollar, and Girshick]{sam}
Alexander Kirillov, Eric Mintun, Nikhila Ravi, Hanzi Mao, Chloe Rolland, Laura Gustafson, Tete Xiao, Spencer Whitehead, Alexander~C. Berg, Wan-Yen Lo, Piotr Dollar, and Ross Girshick.
\newblock Segment anything.
\newblock In \emph{Proceedings of the IEEE/CVF International Conference on Computer Vision}, pages 4015--4026, 2023.

\bibitem[Kobayashi et~al.(2022)Kobayashi, Matsumoto, and Sitzmann]{DFF}
Sosuke Kobayashi, Eiichi Matsumoto, and Vincent Sitzmann.
\newblock Decomposing nerf for editing via feature field distillation.
\newblock In \emph{Proceedings of the Advances in Neural Information Processing Systems}, pages 23311--23330, 2022.

\bibitem[Kopanas et~al.(2021)Kopanas, Philip, Leimk{\"u}hler, and Drettakis]{kopanas2021point}
Georgios Kopanas, Julien Philip, Thomas Leimk{\"u}hler, and George Drettakis.
\newblock Point-based neural rendering with per-view optimization.
\newblock In \emph{Computer Graphics Forum}, pages 29--43. Wiley Online Library, 2021.

\bibitem[Leroy et~al.(2024)Leroy, Cabon, and Revaud]{mast3r}
Vincent Leroy, Yohann Cabon, and J{\'e}r{\^o}me Revaud.
\newblock Grounding image matching in 3d with mast3r.
\newblock In \emph{European Conference on Computer Vision}, pages 71--91, 2024.

\bibitem[Li et~al.(2022)Li, Weinberger, Belongie, Koltun, and Ranftl]{LSeg}
Boyi Li, Kilian~Q Weinberger, Serge Belongie, Vladlen Koltun, and Rene Ranftl.
\newblock Language-driven semantic segmentation.
\newblock In \emph{Proceedings of the International Conference on Learning Representations}, 2022.

\bibitem[Liao and Gao(2024)]{liao2024rethinking}
Guibiao Liao and Wei Gao.
\newblock Rethinking feature mining for light field salient object detection.
\newblock \emph{ACM Transactions on Multimedia Computing, Communications and Applications}, 20\penalty0 (10):\penalty0 1--24, 2024.

\bibitem[Liao et~al.(2020)Liao, Gao, Jiang, Wang, and Li]{liao2020mmnet}
Guibiao Liao, Wei Gao, Qiuping Jiang, Ronggang Wang, and Ge Li.
\newblock Mmnet: Multi-stage and multi-scale fusion network for rgb-d salient object detection.
\newblock In \emph{Proceedings of the 28th ACM international conference on multimedia}, pages 2436--2444, 2020.

\bibitem[Liao et~al.(2022)Liao, Gao, Li, Wang, and Kwong]{liao2022cross}
Guibiao Liao, Wei Gao, Ge Li, Junle Wang, and Sam Kwong.
\newblock Cross-collaborative fusion-encoder network for robust rgb-thermal salient object detection.
\newblock \emph{IEEE Transactions on Circuits and Systems for Video Technology}, 32\penalty0 (11):\penalty0 7646--7661, 2022.

\bibitem[Liao et~al.(2024{\natexlab{a}})Liao, Li, and Ye]{vlm2scene}
Guibiao Liao, Jiankun Li, and Xiaoqing Ye.
\newblock Vlm2scene: Self-supervised image-text-lidar learning with foundation models for autonomous driving scene understanding.
\newblock In \emph{Proceedings of the AAAI Conference on Artificial Intelligence}, pages 3351--3359, 2024{\natexlab{a}}.

\bibitem[Liao et~al.(2024{\natexlab{b}})Liao, Zhou, Bao, Liu, and Li]{ovnerf}
Guibiao Liao, Kaichen Zhou, Zhenyu Bao, Kanglin Liu, and Qing Li.
\newblock Ov-nerf: Open-vocabulary neural radiance fields with vision and language foundation models for 3d semantic understanding.
\newblock \emph{IEEE Transactions on Circuits and Systems for Video Technology}, 2024{\natexlab{b}}.

\bibitem[Liao et~al.(2025)Liao, Li, Bao, Qiu, and Liu]{liao2025spc}
Guibiao Liao, Qing Li, Zhenyu Bao, Guoping Qiu, and Kanglin Liu.
\newblock Spc-gs: Gaussian splatting with semantic-prompt consistency for indoor open-world free-view synthesis from sparse inputs.
\newblock In \emph{Proceedings of the Computer Vision and Pattern Recognition Conference}, pages 11264--11274, 2025.

\bibitem[Liu et~al.(2023{\natexlab{a}})Liu, Lei, Peng, Yu, Li, and Ling]{liu2023novel}
Bingzheng Liu, Jianjun Lei, Bo Peng, Chuanbo Yu, Wanqing Li, and Nam Ling.
\newblock Novel view synthesis from a single unposed image via unsupervised learning.
\newblock \emph{ACM Transactions on Multimedia Computing, Communications and Applications}, 19\penalty0 (6):\penalty0 1--23, 2023{\natexlab{a}}.

\bibitem[Liu et~al.(2022)Liu, Kong, Wang, Li, and Yin]{liu2022spatial}
Caixia Liu, Dehui Kong, Shaofan Wang, Jinghua Li, and Baocai Yin.
\newblock A spatial relationship preserving adversarial network for 3d reconstruction from a single depth view.
\newblock \emph{ACM Transactions on Multimedia Computing, Communications, and Applications}, 18\penalty0 (4):\penalty0 1--22, 2022.

\bibitem[Liu et~al.(2023{\natexlab{b}})Liu, Zhan, Zhang, Xu, Yu, El~Saddik, Theobalt, Xing, and Lu]{3DOVS}
Kunhao Liu, Fangneng Zhan, Jiahui Zhang, Muyu Xu, Yingchen Yu, Abdulmotaleb El~Saddik, Christian Theobalt, Eric Xing, and Shijian Lu.
\newblock Weakly supervised 3d open-vocabulary segmentation.
\newblock In \emph{Proceedings of the Advances in Neural Information Processing Systems}, pages 53433--53456, 2023{\natexlab{b}}.

\bibitem[Liu et~al.(2023{\natexlab{c}})Liu, Zeng, Ren, Li, Zhang, Yang, Jiang, Li, Yang, Su, et~al.]{grounddino}
Shilong Liu, Zhaoyang Zeng, Tianhe Ren, Feng Li, Hao Zhang, Jie Yang, Qing Jiang, Chunyuan Li, Jianwei Yang, Hang Su, et~al.
\newblock Grounding dino: Marrying dino with grounded pre-training for open-set object detection.
\newblock \emph{arXiv preprint arXiv:2303.05499}, 2023{\natexlab{c}}.

\bibitem[Mildenhall et~al.(2020)Mildenhall, Srinivasan, Tancik, Barron, Ramamoorthi, and Ng]{nerf}
Ben Mildenhall, Pratul~P Srinivasan, Matthew Tancik, Jonathan~T Barron, Ravi Ramamoorthi, and Ren Ng.
\newblock Nerf: Representing scenes as neural radiance fields for view synthesis.
\newblock In \emph{Proceedings of the European Conference on Computer Vision}, pages 405--421, 2020.

\bibitem[M{\"u}ller et~al.(2022)M{\"u}ller, Evans, Schied, and Keller]{instantngp}
Thomas M{\"u}ller, Alex Evans, Christoph Schied, and Alexander Keller.
\newblock Instant neural graphics primitives with a multiresolution hash encoding.
\newblock \emph{ACM Transactions on Graphics}, 41\penalty0 (4):\penalty0 1--15, 2022.

\bibitem[Qin et~al.(2024)Qin, Li, Zhou, Wang, and Pfister]{langsplat}
Minghan Qin, Wanhua Li, Jiawei Zhou, Haoqian Wang, and Hanspeter Pfister.
\newblock Langsplat: 3d language gaussian splatting.
\newblock In \emph{Proceedings of the IEEE/CVF Conference on Computer Vision and Pattern Recognition}, pages 20051--20060, 2024.

\bibitem[Radford et~al.(2021)Radford, Kim, Hallacy, Ramesh, Goh, Agarwal, Sastry, Askell, Mishkin, Clark, et~al.]{OpenAICLIP}
Alec Radford, Jong~Wook Kim, Chris Hallacy, Aditya Ramesh, Gabriel Goh, Sandhini Agarwal, Girish Sastry, Amanda Askell, Pamela Mishkin, Jack Clark, et~al.
\newblock Learning transferable visual models from natural language supervision.
\newblock In \emph{Proceedings of the International conference on machine learning}, pages 8748--8763. PMLR, 2021.

\bibitem[Shen et~al.(2024)Shen, Yang, and Wang]{flashsplat}
Qiuhong Shen, Xingyi Yang, and Xinchao Wang.
\newblock Flashsplat: 2d to 3d gaussian splatting segmentation solved optimally.
\newblock In \emph{Proceedings of the European Conference on Computer Vision}, pages 456--472. Springer, 2024.

\bibitem[Shi et~al.(2023)Shi, Wang, Duan, and Guan]{shi2023language}
Jin-Chuan Shi, Miao Wang, Hao-Bin Duan, and Shao-Hua Guan.
\newblock Language embedded 3d gaussians for open-vocabulary scene understanding.
\newblock \emph{arXiv preprint arXiv:2311.18482}, 2023.

\bibitem[Straub et~al.(2019)Straub, Whelan, Ma, Chen, Wijmans, Green, Engel, Mur-Artal, Ren, Verma, et~al.]{replica}
Julian Straub, Thomas Whelan, Lingni Ma, Yufan Chen, Erik Wijmans, Simon Green, Jakob~J Engel, Raul Mur-Artal, Carl Ren, Shobhit Verma, et~al.
\newblock The replica dataset: A digital replica of indoor spaces.
\newblock \emph{arXiv preprint arXiv:1906.05797}, 2019.

\bibitem[Sun et~al.(2025)Sun, Dabral, Zhu, Fua, Theobalt, and Habermann]{sun2024real}
Guoxing Sun, Rishabh Dabral, Heming Zhu, Pascal Fua, Christian Theobalt, and Marc Habermann.
\newblock Real-time free-view human rendering from sparse-view rgb videos using double unprojected textures.
\newblock In \emph{Proceedings of the IEEE/CVF Conference on Computer Vision and Pattern Recognition}, 2025.

\bibitem[Wang et~al.(2024{\natexlab{a}})Wang, Lan, Zhu, Chen, and Lu]{wang2024gsemsplat}
Xingrui Wang, Cuiling Lan, Hanxin Zhu, Zhibo Chen, and Yan Lu.
\newblock Gsemsplat: Generalizable semantic 3d gaussian splatting from uncalibrated image pairs.
\newblock \emph{arXiv preprint arXiv:2412.16932}, 2024{\natexlab{a}}.

\bibitem[Wang et~al.(2024{\natexlab{b}})Wang, Wu, Zhang, and Xu]{wang2024learning}
Yuxin Wang, Qianyi Wu, Guofeng Zhang, and Dan Xu.
\newblock Learning 3d geometry and feature consistent gaussian splatting for object removal.
\newblock In \emph{Proceedings of the European Conference on Computer Vision}, pages 1--17, 2024{\natexlab{b}}.

\bibitem[Wu et~al.(2024{\natexlab{a}})Wu, Bian, Li, Wang, Reid, Torr, and Prisacariu]{wu2024gaussctrl}
Jing Wu, Jia-Wang Bian, Xinghui Li, Guangrun Wang, Ian Reid, Philip Torr, and Victor~Adrian Prisacariu.
\newblock Gaussctrl: Multi-view consistent text-driven 3d gaussian splatting editing.
\newblock In \emph{Proceedings of the European Conference on Computer Vision}, pages 55--71, 2024{\natexlab{a}}.

\bibitem[Wu et~al.(2024{\natexlab{b}})Wu, Zhang, Xu, Jin, Li, Liu, and Loy]{CLIPSelf}
Size Wu, Wenwei Zhang, Lumin Xu, Sheng Jin, Xiangtai Li, Wentao Liu, and Chen~Change Loy.
\newblock Clipself: Vision transformer distills itself for open-vocabulary dense prediction.
\newblock In \emph{Proceedings of the International Conference on Learning Representations}, 2024{\natexlab{b}}.

\bibitem[Yan et~al.(2024)Yan, Lin, Zhou, Wang, Sun, Zhan, Lang, Zhou, and Peng]{streetGaussian}
Yunzhi Yan, Haotong Lin, Chenxu Zhou, Weijie Wang, Haiyang Sun, Kun Zhan, Xianpeng Lang, Xiaowei Zhou, and Sida Peng.
\newblock Street gaussians for modeling dynamic urban scenes.
\newblock In \emph{Proceedings of the European Conference on Computer Vision}, pages 156--173, 2024.

\bibitem[Ye et~al.(2024)Ye, Danelljan, Yu, and Ke]{gaugrouping}
Mingqiao Ye, Martin Danelljan, Fisher Yu, and Lei Ke.
\newblock Gaussian grouping: Segment and edit anything in 3d scenes.
\newblock In \emph{Proceedings of the European Conference on Computer Vision}, pages 162--179. Springer, 2024.

\bibitem[Yu et~al.(2024)Yu, Chen, Huang, Sattler, and Geiger]{mipsplatting}
Zehao Yu, Anpei Chen, Binbin Huang, Torsten Sattler, and Andreas Geiger.
\newblock Mip-splatting: Alias-free 3d gaussian splatting.
\newblock In \emph{Proceedings of the IEEE/CVF Conference on Computer Vision and Pattern Recognition}, pages 19447--19456, 2024.

\bibitem[Zhang et~al.(2023)Zhang, Li, Zou, Liu, Li, Yang, and Zhang]{zhang2023simple}
Hao Zhang, Feng Li, Xueyan Zou, Shilong Liu, Chunyuan Li, Jianwei Yang, and Lei Zhang.
\newblock A simple framework for open-vocabulary segmentation and detection.
\newblock In \emph{Proceedings of the IEEE/CVF International Conference on Computer Vision}, pages 1020--1031, 2023.

\bibitem[Zhang et~al.(2018)Zhang, Isola, Efros, Shechtman, and Wang]{lpips}
Richard Zhang, Phillip Isola, Alexei~A Efros, Eli Shechtman, and Oliver Wang.
\newblock The unreasonable effectiveness of deep features as a perceptual metric.
\newblock In \emph{Proceedings of the IEEE conference on computer vision and pattern recognition}, pages 586--595, 2018.

\bibitem[Zhang et~al.(2024)Zhang, Li, Zhang, Liu, Yuan, Li, Ji, and Lee]{zhang2024street}
Ruida Zhang, Chengxi Li, Chenyangguang Zhang, Xingyu Liu, Haili Yuan, Yanyan Li, Xiangyang Ji, and Gim~Hee Lee.
\newblock Street gaussians without 3d object tracker.
\newblock \emph{arXiv preprint arXiv:2412.05548}, 2024.

\bibitem[Zhang et~al.(2022)Zhang, Liao, Gao, and Li]{zhang2022tdrnet}
Xiaoyu Zhang, Guibiao Liao, Wei Gao, and Ge Li.
\newblock Tdrnet: Transformer-based dual-branch restoration network for geometry based point cloud compression artifacts.
\newblock In \emph{2022 IEEE International Conference on Multimedia and Expo (ICME)}, pages 1--6. IEEE, 2022.

\bibitem[Zhao et~al.(2024)Zhao, Hu, Zhou, Li, and Li]{zhao2024exploiting}
Weichao Zhao, Hezhen Hu, Wengang Zhou, Li Li, and Houqiang Li.
\newblock Exploiting spatial-temporal context for interacting hand reconstruction on monocular rgb video.
\newblock \emph{ACM Transactions on Multimedia Computing, Communications and Applications}, 20\penalty0 (6):\penalty0 1--18, 2024.

\bibitem[Zhou et~al.(2024)Zhou, Chang, Jiang, Fan, Zhu, Xu, Chari, You, Wang, and Kadambi]{feature3dgs}
Shijie Zhou, Haoran Chang, Sicheng Jiang, Zhiwen Fan, Zehao Zhu, Dejia Xu, Pradyumna Chari, Suya You, Zhangyang Wang, and Achuta Kadambi.
\newblock Feature 3dgs: Supercharging 3d gaussian splatting to enable distilled feature fields.
\newblock In \emph{Proceedings of the IEEE/CVF Conference on Computer Vision and Pattern Recognition}, pages 21676--21685, 2024.

\bibitem[Zielonka et~al.(2025)Zielonka, Garbin, Lattas, Kopanas, Gotardo, Beeler, Thies, and Bolkart]{zielonka2025synshot}
Wojciech Zielonka, Stephan~J Garbin, Alexandros Lattas, George Kopanas, Paulo Gotardo, Thabo Beeler, Justus Thies, and Timo Bolkart.
\newblock Synthetic prior for few-shot drivable head avatar inversion.
\newblock In \emph{Proceedings of the IEEE/CVF Conference on Computer Vision and Pattern Recognition}, 2025.

\bibitem[Zuo et~al.(2024)Zuo, Samangouei, Zhou, Di, and Li]{fmgs}
Xingxing Zuo, Pouya Samangouei, Yunwen Zhou, Yan Di, and Mingyang Li.
\newblock Fmgs: Foundation model embedded 3d gaussian splatting for holistic 3d scene understanding.
\newblock \emph{arXiv preprint arXiv:2401.01970}, 2024.

\bibitem[Zwicker et~al.(2001)Zwicker, Pfister, Van~Baar, and Gross]{ewaVolumeSplatting}
Matthias Zwicker, Hanspeter Pfister, Jeroen Van~Baar, and Markus Gross.
\newblock Ewa volume splatting.
\newblock In \emph{Proceedings Visualization, 2001. VIS'01.}, pages 29--538. IEEE, 2001.

\end{thebibliography}
